\documentclass[sigconf]{acmart}

\AtBeginDocument{%
  }

\setcopyright{acmlicensed}
\copyrightyear{2018}
\acmYear{2018}
\acmDOI{XXXXXXX.XXXXXXX}

\acmConference[KDD ’25]{Make sure to enter the correct conference title from your rights confirmation emai}{August 03–07, 2025}{Toronto, ON, Canada}
\acmISBN{978-1-4503-XXXX-X/18/06}

\usepackage[utf8]{inputenc} 
\usepackage[T1]{fontenc}    
\usepackage{hyperref}       
\usepackage{url}            
\usepackage{booktabs}       
\usepackage{amsfonts}       
\usepackage{nicefrac}       
\usepackage{microtype}      
  \usepackage{xcolor}         

\usepackage{amsmath}

\usepackage{graphicx}
\usepackage{caption}
\usepackage{subcaption}
\newcommand{\shortname}{\emph{InvDiff}}

\usepackage{multirow}
\usepackage{array}
\usepackage{wrapfig}
\usepackage{tabularx}
\usepackage{array}
\usepackage{pbox}
\usepackage{bm}
\usepackage{comment}

\usepackage{paralist}

\usepackage{enumitem}

\begin{document}

\title{InvDiff: Invariant Guidance for Bias Mitigation in \\Diffusion Models}

\author{Min Hou}
\authornote{Equal contribution.}
\email{hmhoumin@gmail.com}
\affiliation{%
  \institution{Hefei University of Technology}
  \city{Hefei}
  \country{China}
}

\author{Yueying Wu}
\authornotemark[1]
\email{wuyueying@stu.pku.edu.cn}
\affiliation{%
  \institution{Peking University}
    \city{Beijing}
  \country{China}}

\author{Chang Xu}
\authornote{Corresponding author.}
\email{chanx@microsoft.com}
\affiliation{%
  \institution{Microsoft Research Asia}
    \city{Beijing}
  \country{China}
}

\author{Yu-Hao Huang}
\email{huangyh@smail.nju.edu.cn}
\affiliation{%
  \institution{Nanjing University}
    \city{Nanjing}
  \country{China}}

\author{Chenxi Bai}
\email{chenxibai1149@gmail.com}
\affiliation{%
  \institution{Hefei University of Technology}
  \city{Hefei}
  \country{China}
}

\author{Le Wu}
\email{lewu.ustc@gmail.com}
\affiliation{%
  \institution{Hefei University of Technology}
  \city{Hefei}
  \country{China}
}

\author{Jiang Bian}
\email{jiang.bian@microsoft.com}
\affiliation{%
  \institution{Microsoft Research Asia}
  \city{Beijing}
  \country{China}
}

\renewcommand{\shortauthors}{Trovato et al.}

\begin{abstract}
As one of the most successful generative models, diffusion models have demonstrated remarkable efficacy in synthesizing high-quality images. These models learn the underlying high-dimensional data distribution in an unsupervised manner. Despite their success, diffusion models are highly data-driven and prone to inheriting the imbalances and biases present in real-world data. Some studies have attempted to address these issues by designing text prompts for known biases or using bias labels to construct unbiased data. While these methods have shown improved results, real-world scenarios often contain various unknown biases, and obtaining bias labels is particularly challenging.
In this paper, we emphasize the necessity of mitigating bias in pre-trained diffusion models without relying on auxiliary bias annotations. To tackle this problem, we propose a framework, \shortname, which aims to learn invariant semantic information for diffusion guidance. Specifically, we propose identifying underlying biases in the training data and designing a novel debiasing training objective.
Then, we employ a lightweight trainable module that automatically preserves invariant semantic information and uses it to guide the diffusion model’s sampling process toward unbiased outcomes simultaneously. Notably, we only need to learn a small number of parameters in the lightweight learnable module without altering the pre-trained diffusion model.
Furthermore, we provide a theoretical guarantee that the implementation of \shortname~ is equivalent to reducing the error upper bound of generalization. Extensive experimental results on three publicly available benchmarks demonstrate that \shortname~ effectively reduces biases while maintaining the quality of image generation. Our code is available at https://github.com/Hundredl/InvDiff.

\end{abstract}


\received{20 February 2007}
\received[revised]{12 March 2009}
\received[accepted]{5 June 2009}

\maketitle

\section{INTRODUCTION}
Diffusion models~\cite{ho2020denoising,sohl2015deep,10.5555/3454287.3455354,yang2023diffusion} have emerged as the most successful generative models to date. They have demonstrated remarkable success in synthesizing high-quality images and have also shown potential in a variety of domains, ranging from computer vision~\cite{rombach2022high,song2021scorebased} to temporal data modeling~\cite{fan2024mgtsd,rasul2021autoregressive} and data mining~\cite{wang2024diffcrime}.
The scale of images generated by these models, especially text-to-image diffusion models, is staggering. For instance, over ten million users utilize Stable Diffusion~\cite{rombach2022high} and DALL-E 3~\cite{ramesh2021zero} to generate visually realistic images from textual descriptions~\cite{shen2024finetuning}. Diffusion models learn the underlying high-dimensional data distribution in an unsupervised manner. Despite their success, these models are highly data-driven and prone to inheriting the imbalances and biases present in real-world training data~\cite{10.1145/3649883,kim2024training}. As diffusion models become increasingly prevalent, mitigating the influence of bias becomes more critical, yet this issue has received little attention within the generative model community.

\begin{figure}[!htbp]
    \centering
    \begin{subfigure}[t]{0.49\textwidth}
        \centering
        \includegraphics[width=\textwidth]{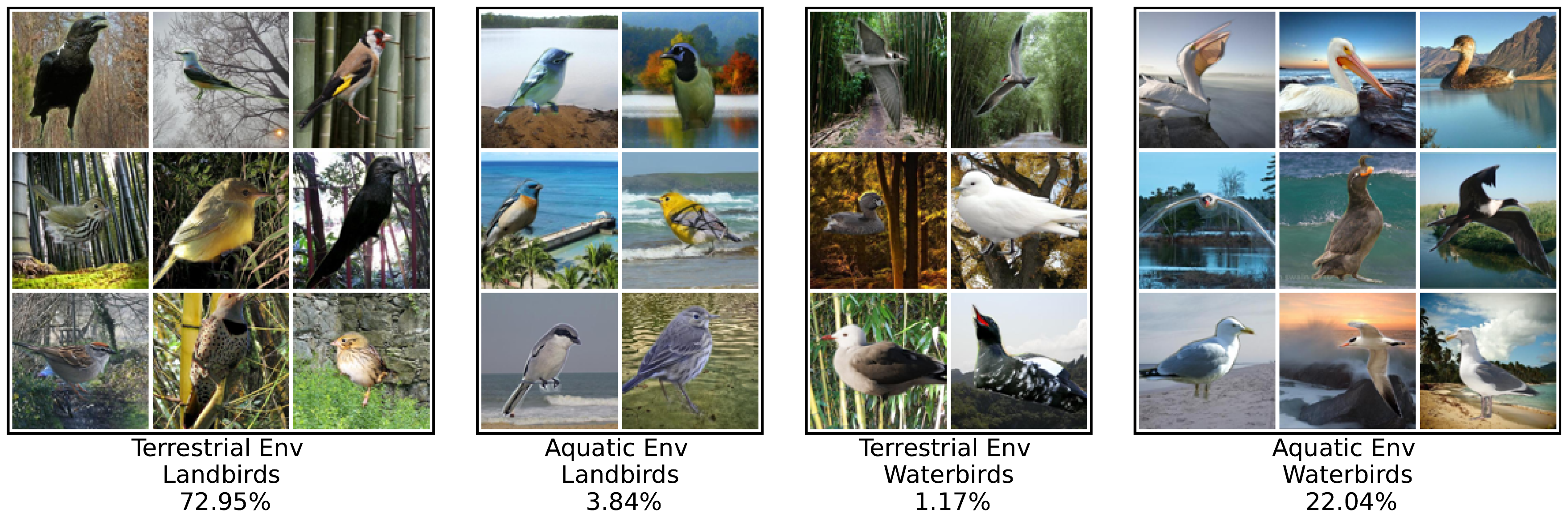}
        \caption{Waterbirds benchmark dataset}
        \label{fig:benchmark}
    \end{subfigure}
    \begin{subfigure}[t]{0.49\textwidth}
        \centering
\includegraphics[width=\textwidth]{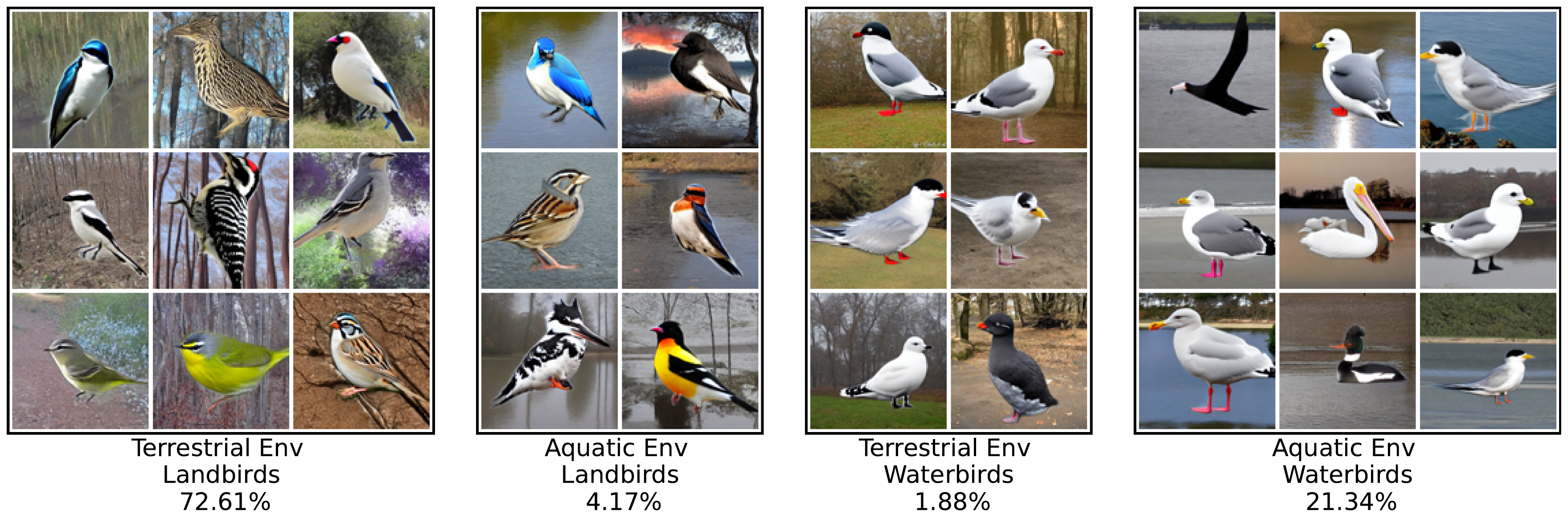}
        \caption{Generated samples from biased model}
        \label{fig:generated}
    \end{subfigure}
    \caption{Proportion and samples of water/land bird in terrestrial/aquatic backgrounds. }
    \label{fig:comparison}
    \Description{}
\end{figure}



Real-world datasets inevitably exhibit biases and undesirable stereotypes, which impact the behavior of diffusion models. To illustrate this, we provide an intuitive experiment demonstrating the impact of biased datasets on the state-of-the-art text-to-image diffusion model, Stable Diffusion. As shown in Figure \ref{fig:comparison}, we fine-tune Stable Diffusion on the biased Waterbirds~\cite{Sagawa*2020Distributionally} dataset (where landbirds are usually in terrestrial backgrounds and waterbirds are usually in aquatic backgrounds). In Figure \ref{fig:comparison}(a), we count the proportions of the two types of birds in the two backgrounds in the original dataset. In Figure \ref{fig:comparison}(b), we report the proportions of the fine-tuned Stable Diffusion model’s generated results, using the types of birds as prompts. We find that the Stable Diffusion model’s generated results unconsciously perpetuate the same bias present in the dataset.
The bias in diffusion models also raises fairness concerns. For example, researchers have found that in images generated by Stable Diffusion, women are underrepresented in high-paying occupations and overrepresented in low-paying ones~\cite{bloomberg2023generative}.
Recently, researchers have made progress in developing diffusion debiasing methods.
Previous efforts can be divided into two categories: (1) Prompt-based methods for known biases~\cite{bansal-etal-2022-well,kim2023stereotyping}. These methods suggest adding ethical interventions to prompts, like “all individuals can be a lawyer irrespective of their gender,” to alleviate specific known biases such as gender and race. (2) Unbiased data-based methods. Given some unbiased data as a prerequisite, these methods train density ratio~\cite{choi2020fair} or a discriminator~\cite{kim2024training} to encourage biased diffusion models convergence to an unbiased distribution.
Despite their success, we argue that these bias annotations such as biases' type and labels are usually unattainable. In fact, real-world data is often complex and contains unknown biases, making these methods less effective when dealing with arbitrary and unknown biases. In such scenarios, obtaining unbiased data becomes even more challenging.

In this work, we investigate the challenging yet practical research problem of \textit{mitigating unknown biases in text-to-image diffusion models without relying on auxiliary bias annotations.} Bias in models often occurs when they learn the spurious correlations~(i.e., shortcut) present in the data~\cite{10.5555/3618408.3620060,9782500}. For instance, if a diffusion model erroneously learns spurious correlations between gender and occupation in training data, it will exhibit bias when generating images from occupation-related text descriptions. Following this line, an intuitive idea is to encourage the diffusion model's sampling process to focus on the semantic information from the text description while neglecting the corresponding spurious correlations. However, due to the entanglement of semantic information and biases, achieving this goal remains challenging. 
To this end, we draw inspiration from invariant learning~\cite{arjovsky2019invariant,creager2021environment}, which can achieve guaranteed performance under distribution shifts and received great attention in recent years. Invariant learning keeps invariant semantic information across different training environments, where environments are variables that should not affect the prediction. 

In this paper, we propose a novel bias mitigation framework \shortname~ for text-to-image diffusion models without relying on auxiliary bias annotation. The main idea of \shortname~ is to encourage the diffusion model to focus on the invariant semantic information from the text description. To preserve the invariant semantic information, we first design a novel debiasing objective. Then we propose a max-min training game for both potential bias annotation inference and bias mitigation. Specifically, we first infer potential bias annotations by maximizing the objective. Given the annotations, we finetune the diffusion model to unbias by minimizing the proposed objective. Notably, we only need to learn a small number of parameters in the lightweight learnable module without altering the pre-trained diffusion model.
Furthermore, we provide a theoretical guarantee that the implementation of \shortname~ is equivalent to reducing the error upper bound of generalization. The main contributions of this work are as follows:
\begin{itemize}[leftmargin=*]
    \item We investigate the challenging yet practical new research problem of mitigating unknown biases in text-to-image diffusion models without relying on auxiliary bias annotations. We propose a novel bias mitigation framework \shortname, which encourages the diffusion model's sampling process to focus on the invariant semantic information.
    \item We design a debiasing objective for diffusion models and propose a max-min training game for both potential bias annotation inference and bias mitigation. We also provide a theoretical guarantee.
    \item Extensive experimental results on three publicly available benchmarks demonstrate that \shortname~ effectively reduces biases while maintaining the quality of image generation.
\end{itemize}

\section{RELATED WORK}
\subsection{Bias in Diffusion Models.}
Over the past few years, 
diffusion models have shown a great ability to generate images with high visual quality. Despite their success, de-biasing is still a fundamental challenge that diffusion models face. Diffusion are known to produce biased and stereotypical images from neutral prompts~\cite{shen2024finetuning}. For instance, researchers~\cite{cho2023dall} found that Stable Diffusion (SD) predominantly produces male images when prompted with various occupations, and the generated skin tone is concentrated on the center few tones. Diffusion models are highly data-driven and prone to inherit bias in real-world data~\cite{10.1145/3600211.3604711,schramowski2023safe}. What's worse, diffusion models not only perpetuate biases
found in the training data but also may amplify it~\cite{seshadri2023bias}.

\subsection{Bias Mitigation in Diffusion Models.}
Recently, researchers have made significant strides in developing debiased diffusion models, primarily focusing on known biases such as genders~\cite{friedrich2023fair,bansal-etal-2022-well}. Most of these efforts are based on prompting techniques. For example, researchers~\cite{friedrich2023fair} proposed adding random text cues like "male" or "female" when specific occupations are detected in prompts to ensure a more balanced gender representation in generated images. Soft prompts~\cite{kim2023stereotyping} are proposed to make doctor images with a balanced gender distribution. Beyond prompting, \cite{shen2024finetuning} suggested extracting face-centric attributes and aligning them with a user-defined target distribution to mitigate gender bias. \cite{kim2024training} addressed debiased diffusion modeling under a weakly supervised setting, requiring some unbiased data as a prerequisite. However, real-world data is often complex and contains unknown biases, making these methods less applicable when dealing with unspecified biases. Moreover, obtaining unbiased data without bias annotations is even more unpractical.
In this work, we introduce \shortname~to tackle the challenging task of bias mitigation without relying on bias annotations.
\vspace{-5pt}
\section{PRELIMINARY}
\subsection{Diffusion Models}
$\bullet$ \textbf{Denoising Diffusion Probabilistic Models.}
Diffusion models~\cite{ho2020denoising} are latent variable models, which aim to model distribution $p_\theta\left(\boldsymbol{x}_0\right)=\int p_\theta\left(\boldsymbol{x}_{0: T}\right) d \boldsymbol{x}_{1: T}$ that approximates the data distribution $\mathbb{P}(\boldsymbol{x}_0)$. Here $\boldsymbol{x}_1, \ldots, \boldsymbol{x}_T$ are latents of the same dimensionality as the data $\boldsymbol{x}_0 \sim \mathbb{P}\left(\boldsymbol{x}_0\right)$. The denoising diffusion models consist of two processes, the forward process and the reverse process, respectively. In the forward process, Gaussian noise is gradually added to the data $\boldsymbol{x}_0$ according to a variance schedule $\left\{\beta_t\right\}_{1: T}$, finally obtaining random noise $\boldsymbol{x}_T$. The process can be formulated as a Markov chain:
\begin{equation}
    \begin{aligned}
        q\left(\boldsymbol{x}_{1: T} \mid \boldsymbol{x}_0\right)&=\prod_{t=1}^T q\left(\boldsymbol{x}_t \mid \boldsymbol{x}_{t-1}\right),
    \\
    \quad q\left(\boldsymbol{x}_t \mid \boldsymbol{x}_{t-1}\right)&=\mathcal{N}\left(\boldsymbol{x}_t ; \sqrt{1-\beta_t} \boldsymbol{x}_{t-1}, \beta_t \mathbf{I}\right).
    \end{aligned}
\end{equation}
The noisy distribution at any intermediate timestep is $q\left(\boldsymbol{x}_t \mid \boldsymbol{x}_{0}\right)=\mathcal{N}\left(\boldsymbol{x}_t; \sqrt{\bar{\alpha}_t} \boldsymbol{x}_0,\left(1-\bar{\alpha}_t\right) \mathbf{I}\right)$, and $\bar{\alpha}_t=\prod_{i=1}^t\left(1-\beta_i\right)$. Namely, $\boldsymbol{x}_t=\sqrt{\bar{\alpha}_t} \boldsymbol{x}_0+\sqrt{1-\bar{\alpha}_t} \boldsymbol{\epsilon}$, where $\boldsymbol{\epsilon} \sim \mathcal{N}\left(\mathbf{0}, \mathbf{I}\right)$ is a Gaussian noise. 

In the reverse process, a generative model $\theta$ learns to estimate the analytical true posterior in order to gradually recover $\boldsymbol{x}_{0}$ from a Gaussian noise input  $\boldsymbol{x}_{T} \sim \mathcal{N}\left(\mathbf{0}, \mathbf{I}\right)$. The process can be defined as a Markov chain:
\begin{equation}
\begin{aligned}
    p_\theta\left(\boldsymbol{x}_{0: T}\right)&=p\left(\boldsymbol{x}_T\right) \prod_{t=1}^T p_\theta\left(\boldsymbol{x}_{t-1} \mid \boldsymbol{x}_t\right), 
    \\
    \quad p_\theta\left(\boldsymbol{x}_{t-1} \mid \boldsymbol{x}_t\right)&=\mathcal{N}\left(\boldsymbol{x}_{t-1} ; \mu_\theta\left(\boldsymbol{x}_t, t\right), \Sigma_\theta\left(\boldsymbol{x}_t, t\right) \mathbf{I}\right).
\label{eq:reverse}
\end{aligned}
\end{equation}
The optimization objective is to minimize the KL divergence between 
$q\left(\boldsymbol{x}_{t-1} \mid \boldsymbol{x}_t, \boldsymbol{x}_0\right)$ and $  p_\theta\left(\boldsymbol{x}_{t-1} \mid \boldsymbol{x}_t\right)$. According to DDPM~\cite{ho2020denoising}, the parameterization of $p_\theta\left(\boldsymbol{x}_{t-1} \mid \boldsymbol{x}_t\right)$ is chosen as:
\begin{equation}
\mu_\theta\left(\boldsymbol{x}_t, t\right)=\frac{1}{\sqrt{\alpha_n}}\left(\boldsymbol{x}_t-\frac{1-\alpha_t}{\sqrt{1-\bar{\alpha}_t}} \boldsymbol{\epsilon}_\theta\left(\sqrt{\bar{\alpha}_t} \boldsymbol{x}_0+\sqrt{1-\bar{\alpha}_t} \boldsymbol{\epsilon}, t\right)\right),
\end{equation}
where $\boldsymbol{\epsilon}_\theta(\boldsymbol{x}_t,t)$ is a neural network which predicts the Gaussian noise $\boldsymbol{\epsilon}$. The variance $\Sigma_\theta\left(\boldsymbol{x}_t, t\right)$ can be fixed to untrained time-dependent constants.
The objective can be reduced to a simple denoising estimation loss:
\begin{equation}
\mathcal{L}_{\mathrm{DDPM}}=\mathbb{E}_{t, \boldsymbol{x}_0 \sim q\left(\boldsymbol{x}_0\right), \boldsymbol{\epsilon} \sim \mathcal{N}(\mathbf{0}, \mathbf{I})}\left[\left\|\boldsymbol{\epsilon}-\boldsymbol{\epsilon}_\theta\left(\sqrt{\bar{\alpha}} \boldsymbol{x}_0+\sqrt{1-\overline{\alpha_t}} \boldsymbol{\epsilon}, t\right)\right\|^2\right].
\end{equation}
 After training the $\boldsymbol{\epsilon}_\theta$ and given a Gaussian noise input, we can iteratively sample from the reverse process to reconstruct $\boldsymbol{x}_0$.

\noindent$\bullet$ \textbf{Text-to-Image Diffusion Models.}
In cases where text description (i.e., prompts)~\cite{liu2023more} $\boldsymbol{y}$ are available, diffusion models can model conditional distributions of the form $p(\boldsymbol{x}_t | \boldsymbol{y})$. 
Some studies, including Stable Diffusion~\cite{rombach2022high}, implement conditional diffusion with a conditional denoising autoencoder $\boldsymbol{\epsilon}_\theta\left(\boldsymbol{x}_t, t, \boldsymbol{y}\right)$ and paves the way to controlling the synthesis process through input text prompt $\boldsymbol{y}$. In the sampling, the label-guided model estimates the noise with a linear interpolation $\hat{\boldsymbol{\epsilon}}=(1+\omega) \boldsymbol{\epsilon}_\theta\left(\boldsymbol{x}_t, t, \boldsymbol{y}\right)-\omega \boldsymbol{\epsilon}_\theta\left(\boldsymbol{x}_t, t\right)$ to recover $\boldsymbol{x}_{t-1}$, which is often referred as classifier-free guidance~\cite{ho2021classifierfree}.

\subsection{Invariant Learning}
\label{sec:invariant_learning}

\noindent $\bullet$ \textbf{The Environment Invariance Constraint~(EIC).} 
Invariant learning~(IL)~\cite{arjovsky2019invariant} is an emerging technique for improving discriminative models' robustness by blocking spurious correlations in data. IL is based on the assumption that the causal mechanism remains invariant across various environments~(a.k.a., domains), while the spurious correlation varies. For example, the correlation between the background green grass and the label landbird is unstable across the images of data collected from different locations.  In this light, IL pushes models to capture the causal mechanism by penalizing the variance of model performance across environments. 

Formally, consider the task of learning a predictor $f: \mathcal{X} \rightarrow \mathcal{Y}$, which maps input $x \in \mathcal{X}$ to output $y \in \mathcal{Y}$. Suppose the predictor $f$ can be decomposed into $f=\omega \circ \Phi$, where $\Phi: \mathcal{X} \rightarrow \mathcal{H}$ denotes a feature encoder which maps the input into a representation space $\mathcal{H}$, $\omega: \mathcal{H} \rightarrow \mathcal{Y}$ is a classifier. Suppose the training data $\mathcal{D}$ are collected under multiple environments $\mathcal{E}$, i.e., $\mathcal{D}=\left\{D_e\right\}_{e \in \mathcal{E}}$. $D_e=\left\{x_i^e, y_i^e\right\}_{i=1}^{n^e}$ contains data sampled from the probability distribution $\mathbb{P}^e(\mathcal{X} \times \mathcal{Y})$.
The target of IL is to encourage the encoder $\Phi$ to extract invariant features associated with causal mechanisms by satisfying the following constraint:
\begin{equation}
\mathbb{P}(Y \mid \Phi(X), E=e)=\mathbb{P}\left(Y \mid \Phi(X), E=e^{\prime}\right), \forall e, e^{\prime} \in \mathcal{E}.
\label{eq:eic}
\end{equation}
The constraint is termed as \textit{Environment Invariance Constraint~(EIC).} The constraint can be incorporated into the training target via a penalty term. The learning object of IL can be written as~\cite{krueger2021out}:
\begin{equation}
\min _{\omega, \Phi} \sum^e \mathcal{R}^e(\omega, \Phi)+\lambda \operatorname{Var}\left(\mathcal{R}^e(\omega, \Phi)\right),
\label{eq:invariant_loss}
\end{equation}
where the $\mathcal{R}^e(\omega, \Phi)$ represents the training loss of $f$ on the environment $e$, and the second term is the constraint over the variance across environments. The training process of Eq.(\ref{eq:invariant_loss}) enforces the optimal classifier $\omega^*$ on top of the representation space to be the same across all environments, therefore encouraging the encoder $\Phi$ extract invariant and stable features automatically.

Note that invariant learning was originally developed for discriminative models. This approach requires learning an encoder $\Phi$ to extract features, and then based on these features performing a classification task to ensure balanced performance across different environments. However, this method cannot be directly applied to diffusion models. As in diffusion, $x$ is generated rather than provided, making it infeasible to extract invariant features of $x$. Additionally, there is no classifier $\omega$ mapping feature embedding to $y$ for generative tasks. In this paper, inspired by invariant learning, we designed a debiasing objective applicable to diffusion models.

\noindent $\bullet$ \textbf{Inviariant Learning without Bias Annotation.} Recently, invariant learning has been extended to the scenario where environment labels~(i.e., bias annotations) are unknown~\cite{creager2021environment,chen2022when}. These methods utilize prior knowledge of spurious correlations to divide the training data into groups. A notable work is EIIL~\cite{creager2021environment}, which inferences environments and groups training data by maximizing violations of the EIC principle. Specifically, EIIL splits the training data into groups such that the label distribution conditioned on the spurious feature varies maximally. Similar to predefined environments, these groups are intended to encode variations of spurious information while preserving the causal mechanism.

\section{METHODOLOGY}
In this section, we present our \shortname~framework, which finetunes biased diffusion models to unbiased without bias annotation. We start by formulating the debiasing target in Section \ref{sec:problem_formulation}. 
Section \ref{sec:invariant_guidence} explains the motivation behind invariant guidance.
Subsequently, in Section \ref{sec:invdiff_irm}, we introduce the proposed debiasing objective for both potential bias annotation inference and bias mitigation. We provide a theoretical analysis in Section \ref{sec:theoretical}.

\begin{figure*}[ht]
\centering
\includegraphics[width = 1\textwidth]{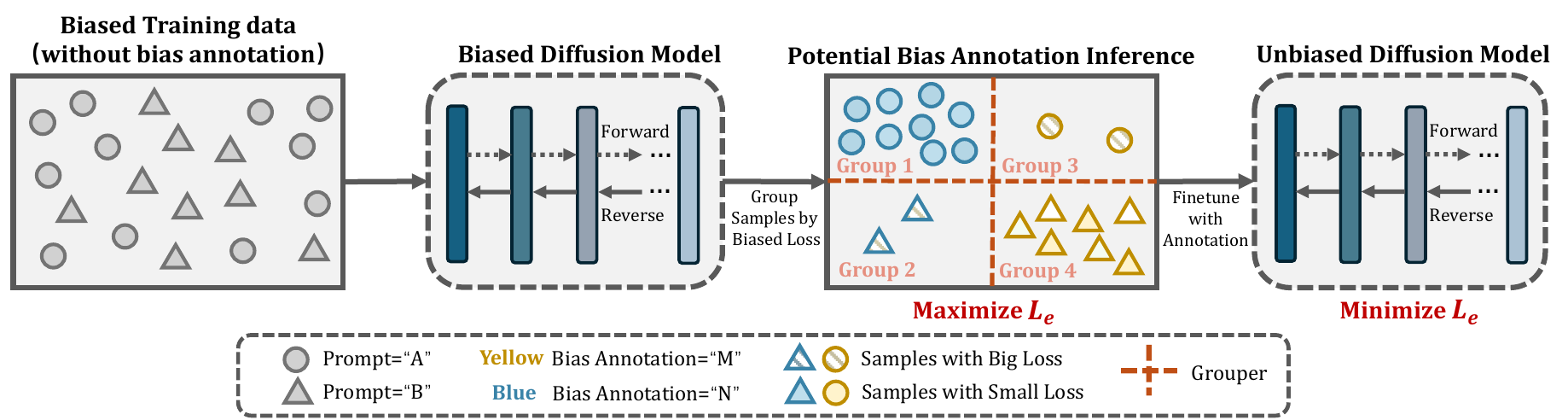}
\caption{An overview of \shortname. We first design a novel debiasing objective $\mathcal{L}_e$ for diffusion models. Then we propose a max-min game with the debiasing objective. We first infer potential bias annotations by maximizing the objective. Given the annotations, we finetune the biased model to unbiased by minimizing the proposed objective.}
\label{pic:framework}
\end{figure*}

\subsection{Formalization of Unbiased Diffusion Model}
\label{sec:problem_formulation}
Given training dataset $\mathcal{D}=\{x_i,y_i\}_{i=1}^N$, data $x \in \mathcal{X}$ and text prompts $y \in \mathcal{Y}$, the training process of current text-to-image diffusion models is to try its best to approximate the conditional distribution $\mathbb{P}(X|Y)$ in training data. However, real-world data often contain spurious correlations, which are correlations between meaningless features and text prompts. Diffusion models are prone to learning the easy-to-fit spurious correlations, resulting in generating biased data. For example, if most of the presidents in the training set are men, the diffusion model may learn about the spurious correlations between the job and the gender. When generating images taking "president" as a text prompt, the model is highly likely to generate a male president.
We can denote $\boldsymbol{x}^{inv}_y$ as the \textit{invariant semantic information} of an instance $\boldsymbol{x}$ which defines its text prompt $\boldsymbol{y}$~(e.g., the semantic information of a person in political scenes with formal attire), where $\boldsymbol{x}^{sp}_y$ as spurious correlation toward $\boldsymbol{y}$~(e.g., gender characteristics). $X^{inv}$ and $X^{sp}$ denote the corresponding random variables.
In this work, our goal is to mitigate bias in diffusion models by eliminating the influence of arbitrary and unknown spurious correlations. Namely, we aim to obtain a debiasing diffusion model whose conditional generation results only depend on $\mathbb{P}(X^{inv}|Y)$.

\subsection{Invariant Guidance}
\label{sec:invariant_guidence}
Generally, one can have a pre-trained biased text-to-image diffusion model on dataset $\mathcal{D}$, e.g., DDPM model with parameters $\theta$,
$p_\theta\left(\boldsymbol{x}_{t-1} \mid \boldsymbol{x}_t, \boldsymbol{y}\right)=\mathcal{N}\left(\boldsymbol{x}_{t-1}; \mu_\theta\left(\boldsymbol{x}_t, t, \boldsymbol{y}\right), \sigma_t \mathbf{I}\right)$, which are trained to fit the biased conditional distribution $\mathbb{P}(X|Y)$ on $\mathcal{D}$.
We denote the parameters of the ideal text-to-image diffusion model whose conditional generation results only depend on $\mathbb{P}(X^{inv}|Y)$ as
$p_\phi\left(\boldsymbol{x}_{t-1} \mid \boldsymbol{x}_t,\boldsymbol{y}\right)=\mathcal{N}\left(\boldsymbol{x}_{t-1}; \mu_\phi\left(\boldsymbol{x}_t, t, \boldsymbol{y}\right), \sigma_t \mathbf{I}\right)$ with parameters $\phi$.
The spurious correlation between $\boldsymbol{y}$ and $\boldsymbol{x}$ leads to the generated results of pre-trained diffusion models containing $\boldsymbol{x}^{sp}_y$ (e.g., when taking "president" as a condition, the model always generate a male president), while the ideal diffusion model can generate samples only depend on $\boldsymbol{x}^{inv}_y$ (e.g., generate president images with various genders).
Due to the existence of spurious correlations, there is a gap between the posterior mean predicted by the actual conditional diffusion model~($\mu_\theta\left(\boldsymbol{x}_t, t, \boldsymbol{y}\right)$) and the ideal one~($\mu_\phi\left(\boldsymbol{x}_t, t, \boldsymbol{y}\right)$). 

From the perspective of the posterior mean gap, we can draw inspiration from the classifier-guided sampling method~\cite{dhariwal2021diffusion,zhang2022unsupervised,yang2023disdiff}. The classifier-guided sampling methods show that one can train a classifier $p_\phi\left(\boldsymbol{y} \mid \boldsymbol{x}_t\right)$ and use its gradient $\nabla_{\boldsymbol{x}_t} \log p_\phi\left(\boldsymbol{y} \mid \boldsymbol{x}_t\right)$ as the mean shift item to guide pre-trained unconditional diffusion model to sample towards specified class $\boldsymbol{y}$. By introducing the prior knowledge from $\boldsymbol{y}$, the classifier-guided sampling methods fill the posterior mean gap between the unconditional diffusion process and the ideal conditional diffusion process.
Similar to the classifier-guided sampling method that utilizes condition $\boldsymbol{y}$ as prior knowledge to fill the gap,
we aim to introduce invariant features $\boldsymbol{x}^{inv}_y$ as prior knowledge for biased pre-trained conditional diffusion model. The mean shift item from invariant features can help the reverse process fill the gap and focus on reconstructing invariant features rather than spurious correlations. \shortname~follows this principle to obtain an unbiased diffusion model based on a pre-trained biased diffusion model.

Given a pre-trained biased text-to-image diffusion model on dataset $\mathcal{D}$, $p_\theta\left(\boldsymbol{x}_{t-1} \mid \boldsymbol{x}_t, \boldsymbol{y}\right)=\mathcal{N}\left(\boldsymbol{x}_{t-1}; \mu_\theta\left(\boldsymbol{x}_t, t, \boldsymbol{y}\right), \sigma_t \mathbf{I}\right)$, our target is to utilize the knowledge from invariant semantic information to guide the diffusion process. Similar to the classifier-guided sampling methods that use the gradient $\nabla_{\boldsymbol{x}_t} \log p\left(\boldsymbol{y} \mid \boldsymbol{x}_t\right)$ as the mean shift item to guide pre-trained unconditional diffusion model to sample towards specified condition $\boldsymbol{y}$, we use the gradient information from $\boldsymbol{x}^{inv}_y$, i.e., $\nabla_{\boldsymbol{x}_t} \log p\left(\boldsymbol{x}^{inv}_y \mid \boldsymbol{x}_t \right)$, making the diffusion process more focuses on invariant semantic information.
Therefore, the target diffusion process can be formulated as:
\begin{equation}
\begin{aligned}
p_{\phi}&\left(\boldsymbol{x}_{t-1} \mid \boldsymbol{x}_t, \boldsymbol{y}\right) \approx \\& \mathcal{N}\left(\boldsymbol{x}_{t-1} ; \boldsymbol{\mu}_\theta\left(\boldsymbol{x}_t, t, \boldsymbol{y}\right)+\sigma_t \cdot \nabla_{\boldsymbol{x}_t} \log p\left(\boldsymbol{x}^{inv}_y \mid \boldsymbol{x}_t\right), \sigma_t\mathbf{I}\right).
\end{aligned}
\label{eq:sampling_process}
\end{equation}
If $\boldsymbol{x}^{inv}_y$ is available, we can obtain the $\nabla_{\boldsymbol{x}_t} \log p_\phi\left(\boldsymbol{x}^{inv}_y \mid \boldsymbol{x}_t\right)$ and directly utilize the Eq.(\ref{eq:sampling_process}) to guide the sampling process. Nevertheless, obtaining the invariant semantic information is not a trivial task. Specifically, (\textit{i}) $\boldsymbol{x}^{inv}_y$ is generally not directly provided by the data set. (\textit{ii}) The extraction of $\boldsymbol{x}^{inv}_y$ does not follow a unified rule. For instance, in the bird generation task, the foreground serves as the invariant information, whereas in the grassland generation task, the background is the invariant information. So we can't directly extract fixed foreground or background et al., as semantic information. 
Therefore, there is still an urgent need to design a more general data-driven approach to preserve invariant representations.
In the next subsection, we will elaborate on our solution.

\subsection{\shortname~with Invariant Semantic Information Learning}
\label{sec:invdiff_irm}
In this subsection, we introduce the detailed invariant semantic information-preserving method. We first design a novel debiasing objective for diffusion models. Then we propose a max-min game with the objective. We first infer potential bias annotations by maximizing the objective. Given the annotations, we finetune the biased model to unbiased by minimizing the proposed objective. The overview of \shortname~is shown in Figure \ref{pic:framework}.

\noindent$\bullet$ \textbf{Debiasing Objective for Diffusion Models.}
Inspired by invariant learning~\cite{creager2021environment} as discussed in Section \ref{sec:invariant_learning}, which using the \textit{Environment Invariance Constraint (EIC)} (Eq. \ref{eq:eic}) to encourage the encoder $\Phi$ to extract
invariant features. 
We propose incorporating this approach into diffusion models to automatically learn the invariant representation $\boldsymbol{x}^{inv}_y$. In general, the training process of \shortname~generally comprises two phases: \textit{(i) Potential Bias Annotation Inference} and \textit{(ii) Invariant Learning Regularization}. In phase \textit{(i)}, training data is grouped into environments by maximizing violations of the EIC principle. These groups intend to encode variations of spurious information while preserving the causal mechanism. The phase \textit{(ii)} employs EIC as a regularization term to learn invariant representations based on the grouping results from the previous phase. We will elaborate on them in detail.

\noindent$\bullet$ \textbf{Potential Bias Annotation Inference.} In this subsection, we introduce a novel differentiable bias annotation inference method for diffusion models.
We maximize violation of the EIC principle to divide the training data into several groups~(i.e., environments), here the groups are expected to hold the invariant mechanism and reflect spurious correlations.
Specifically, we use a learnable matrix $\mathbf{W} \in \mathbb{R}^{N \times E}$ to indicate which group the sample belongs to. Here $N$ is the number of training samples and $E$ is a hyperparameter representing the number of groups. $\mathbf{W}_{ne}$ represent the probability of sample $n$ belongs to group $e$, i.e., $\sum_e \mathbf{W}_{ne} = 1$ and $\mathbf{W}_{ne} \geq 0$. We optimize $\mathbf{W}$ with the following objectives:
\begin{align}
\label{eq:grouper_loss}
\mathbf{W}^* &= \arg \max _{\mathbf{W}} \left(\mathop{Var}\limits_{e} (\mathcal{L}_e) + \omega \min\limits_{e}(\mathcal{L}_e) \right), \\
\mathcal{L}_e &= \frac{1}{N_e} \sum_n^N \mathbf{W}_{ne} \left\|\boldsymbol{\epsilon}-\boldsymbol{\epsilon}_\theta\left(\boldsymbol{x}^n_{t}, t, \boldsymbol{y}^n\right)\right\|^2.
\end{align}
We define $\mathcal{L}_e$ as the diffusion loss within environment $e$, while $\mathop{Var}(\cdot)$ denotes the variance calculation. $\boldsymbol{\epsilon}_\theta$ is the pre-trained biased diffusion model, which remains fixed. 
The regularization term $\omega \min\limits_{e}(\mathcal{L}_e)$ ensures every group contains samples. $\omega$ is the hyperparameter of dispersion
degree.
By maximizing the variance of loss across different environments, we obtain the group indicator matrix $\mathbf{W}$.
We can use the sample grouping result vector $\mathbf{W}_n \in \mathbb{R}^E$ as bias annotation. Each value in the vector represents the probability of the sample $n$ belonging to the corresponding group $e$.

\noindent $\bullet$ \textbf{Invariant Learning Regularization.}
After grouping training data into environments, in the invariant learning regularization phase we employ EIC as a regularization term. We learn invariant representations based on the grouping results and the minimization of EIC. 
Note that as discussed in Section \ref{sec:invariant_learning}, invariant learning cannot be directly applied to diffusion models. As $\boldsymbol{x}$ are generated rather than given, making it impossible to extract invariant features from $\boldsymbol{x}$. Furthermore, there is no classifier to map feature embeddings to label in generative tasks.
To address these challenges specific to generative tasks, we shift our focus to extracting features from $\boldsymbol{y}$, replacing the $\boldsymbol{x}$ encoder to capture invariant information effectively.
Specifically, we employ an encoder $\Phi(\boldsymbol{y})$ for learning invariant representations. Then, we incorporate $\Phi(\boldsymbol{y})$ and the EIC regularization term into Eq.(\ref{eq:sampling_process}), and rewrite it as:
\begin{equation}
\small
\begin{aligned}
\mathcal{L}_{\psi,\Phi}&=\underset{\boldsymbol{x}_0, t, \epsilon}{\mathbb{E}}\left[\left\|\epsilon-\boldsymbol{\epsilon}_\theta\left(\boldsymbol{x}_t, t,\boldsymbol{y}\right)+ \Delta \boldsymbol{G}_\psi\left(\boldsymbol{x}_t, \Phi
(\boldsymbol{y}), t\right)\right\|^2\right] \\
&+ \lambda \mathop{Var}\limits_{e}(\frac{1}{N_e} \sum_n^N \mathbf{W}^*_{ne} \left\|\boldsymbol{\epsilon}-\boldsymbol{\epsilon}_\theta\left(\boldsymbol{x}^n_{t}, t, \boldsymbol{y}^n\right) + \Delta\boldsymbol{G}_\psi\left(\boldsymbol{x}^n_t, \Phi
(\boldsymbol{y}^n), t\right)\right\|^2).
\end{aligned}
\label{eq:final_loss}
\end{equation}
Since $\nabla_{\boldsymbol{x}_t} \log p\left(\boldsymbol{x}^{inv}_y \mid \boldsymbol{x}_t\right)$ is intractable,
we can employ a gradient estimator $\boldsymbol{G}_\psi\left(\boldsymbol{x}_t, \Phi(\boldsymbol{y}), t\right)$ to simulate it.
$\lambda$ and $\Delta$ is the hyper-parameters. The pretrained diffusion model $\theta$ and the group indicate matrix $\mathbf{W}$ is fixed. The EIC regularization term encourages the encoder $\Phi$ to extract invariant semantic information automatically.
We found that a lightweight module $\boldsymbol{G}_\psi$ with only a small number of parameters can yield effective results without altering the pre-training diffusion model. 
$\boldsymbol{G}_\psi$ serves as a mean shift item to only guide diffusion model's sampling process more focuses on invariant information. Compared to fine-tuning the entire model, it is easier to find optimal solutions for debiasing while maintaining generation quality.

\subsection{Theoretical Analysis}
\label{sec:theoretical}
The cause of bias is that the model incorrectly learns the spurious correlation
Similarly, generalization issues occur when models rely on spurious correlation, resulting in poor performance on new data. 
Therefore, the bias mitigation problem can be viewed as a special case of the generalization problem.
To this end, we conducted analyses from the perspective of out-of-distribution generalization to demonstrate that \shortname~ is theoretically supported.
Our analysis demonstrates that implementing \shortname effectively reduces the error upper bound of generalization, thereby proving its effectiveness.
Drawing inspiration from prior research \cite{sicilia2023domain,10402053}, we have the following proposition.

\textbf{Proposition 1.} 
\textit{(Proposition 2.1 in \cite{sicilia2023domain})}
\textit{Let $\mathcal{X}$ be a space, $\mathcal{H}$ be a class of hypotheses corresponding to this space, and $d_{\mathcal{H}\Delta \mathcal{H}}$ be the $\mathcal{H}$-divergence that measures distributional differences. 
Let $\mathbb{Q}$ be the target distribution and the collection $\{\mathbb{P}_i\}^k_{i=1}$ be distributions over $\mathcal{X}$ and let $\{\varphi_i\}^k_{i=1}$ be a collection of non-negative coefficients with $\sum_i \varphi_i =1$. 
Let $\mathcal{O}$ be a set of distributions such that for every $\mathbb{S}\in \mathcal{O}$ the following holds:}
\begin{equation}\label{eq13}
    \sum_i \varphi_i d_{\mathcal{H}\Delta \mathcal{H}} (\mathbb{P}_i,\mathbb{S})\leq \max_{i,j} d_{\mathcal{H}\Delta \mathcal{H}} (\mathbb{P}_i,\mathbb{P}_j).
\end{equation}
Then, for any $h\in \mathcal{H}$, the error on the target domain $\mathbb{Q}$, denoted as $\varepsilon_\mathbb{Q}(h)$, is proven to satisfie the following error upper bound\cite{sicilia2023domain}:
\begin{equation}\label{eq14}
\small
\begin{split}
    \varepsilon_\mathbb{Q}(h)\leq
    \underbrace{\lambda_\varphi}_{\uppercase\expandafter{\romannumeral1}} + 
    \underbrace{\sum_i \varphi_i \varepsilon_{\mathbb{P}_{i}}(h)}_{\uppercase\expandafter{\romannumeral2}} +
    \underbrace{\frac{1}{2} \min_{\mathbb{S}\in \mathcal{O}} d_{\mathcal{H}\Delta \mathcal{H}} (\mathbb{S},\mathbb{Q})} _{\uppercase\expandafter{\romannumeral3}}\\ +
    \underbrace{\frac{1}{2} \max_{i,j} d_{\mathcal{H}\Delta \mathcal{H}} (\mathbb{P}_i,\mathbb{P}_j)}_{\uppercase\expandafter{\romannumeral4}},
\end{split}
\end{equation}
\textit{where $\lambda_\varphi=\sum_i \varphi_i \lambda_i$ and each $\lambda_i$ is the error of an ideal joint hypothesis for  $\mathbb{Q}$ and $\mathbb{P}_i$, $\varepsilon_{\mathbb{P}_i}(h)$ is the error for a hypothesis $h$ on distribution $\mathbb{P}_i$.
}

From Proposition 1, the upper bound of the model's error in the unseen target domain $\mathbb{Q}$ can be expressed as Eq. \eqref{eq14}.
A lower value of $\varepsilon_\mathbb{Q}(h)$ indicates better generalization performance of the model.
Then, we analyze each term of Eq. \eqref{eq14}.
For term $\uppercase\expandafter{\romannumeral1}$, $\lambda_\varphi$ can be ignored in practice because it is small in reality. 
For term $\uppercase\expandafter{\romannumeral2}$, $\sum_i \varphi_i \varepsilon_{\mathbb{P}{i}}(h)$ represents the error in the training domain.
Empirical Risk Minimization (ERM) is an appropriate method for controlling this term.
\shortname optimizes it by minimizing the first term in Eq.(\ref{eq:final_loss}).
For term $\uppercase\expandafter{\romannumeral3}$, $\frac{1}{2} \min_{\mathbb{S}\in \mathcal{O}} d_{\mathcal{H}\Delta \mathcal{H}}(\mathbb{S},\mathbb{Q})$ is the smallest  $\mathcal{H}$-divergence between $\mathbb{S}$ and $\mathbb{Q}$.
Given that $\mathbb{Q}$ is unknown, the only way to reduce this term is to expand the range of $\mathcal{O}$, thereby increasing the likelihood of finding an $\mathbb{S}$ that is closer to $\mathbb{Q}$.
According to Eq. \eqref{eq13}, maximizing the distribution gap between $\mathbb{P}_i$ and $\mathbb{P}_j$ achieves this.
In \shortname, we infer group labels by maximizing $\mathcal{L}_{e}$, which increases the distributional disparity between groups.
For term $\uppercase\expandafter{\romannumeral4}$, $\frac{1}{2} \max_{i,j} d_{\mathcal{H}\Delta \mathcal{H}}(\mathbb{P}_i,\mathbb{P}_j)$ represents the maximum pairwise $\mathcal{H}$-divergence among the source domains.
\shortname~ minimizes $\mathcal{L}_{e}$ to reduce the differences between different domains, thereby decreasing the value of $\frac{1}{2} \max_{i,j} d_{\mathcal{H}\Delta \mathcal{H}}(\mathbb{P}_i,\mathbb{P}_j)$.

\section{EXPERIMENTS}

\begin{table*}[!h]
\small
    \caption{Overall Image Generation Results.}
    \label{tab:tab_table1}
    \centering
    \begin{tabular}{c|cccc|cccc|cccc}
    \toprule
     & \multicolumn{4}{c|}{Waterbirds} & \multicolumn{4}{c|}{CelebA} &  \multicolumn{4}{c}{FairFace}\\  
     \midrule
     Model& FID $\downarrow$& Rec $\uparrow$& Bias$\downarrow$& CLIP-T $\uparrow$& FID $\downarrow$& Rec $\uparrow$& Bias$\downarrow$& CLIP-T $\uparrow$ & FID $\downarrow$& Rec $\uparrow$& Bias$\downarrow$&CLIP-T $\uparrow$\\ 
     \midrule
     Stable Diffusion& 78.13& 0.16& 0.89(0.18)& 31.43(2.22)& 70.52& 0.49& 0.80(0.01)& 26.66(2.16)& 115.90& 0.31& 0.41(0.02)&26.04(1.24)\\
     \midrule
 TIW& 156.90& 0.00& 0.87(0.22)& 21.15(1.96)& 91.63& 0.02& 0.49(0.29)& 24.03(0.85)& 67.31& 0.18& 0.36(0.01)&26.90(1.32)\\
 Fair-Diffusion& — & — & — & —& 199.09 & 0.50 &  0.46(0.15)&27.82(1.32) &229.50 & 0.40& 0.15(0.06)&26.61(1.31)\\
 \midrule
 \shortname-Full-Hard& 71.03& 0.33& 0.69(0.33)& 30.47(2.84)& 76.78& 0.60& 0.41(0.24)& 26.85(2.42)& 126.45& 0.49& 0.20(0.04)&25.99(1.25)\\
 \shortname-Part-Hard& 78.08& 0.16& 0.69(0.39)& 31.58(2.23)& 85.17& 0.59& 0.34(0.17)& 26.38(2.70)& 120.39& 0.57& 0.11(0.01)&26.26(1.28)\\
 \shortname-Full-Soft& 84.27& 0.22& 0.86(0.21)& 29.05(2.87)& 75.31& 0.50& 0.75(0.12)& 26.65(2.25)& 121.07& 0.41& 0.21(0.14)&25.91(1.27)\\ 
     \shortname-Part-Soft& 79.85&0.18& 0.66(0.33)& 33.22(2.01)& 105.2& 0.54& 0.70(0.16)& 26.86(2.68)& 94.01& 0.62& 0.22(0.01)&26.21(1.29)\\ 
     \bottomrule
    \end{tabular}
\label{tab:tab_main}
\end{table*}
We conduct experiments to answer the following questions:
\begin{itemize}[leftmargin=*]
\item \textbf{RQ1:} Can our \shortname~ mitigate biases in generated images and maintain quality under various experimental settings?
\item \textbf{RQ2:} How do some important designs and hyperparameters affect the model? What is the time and space complexity?
\item \textbf{RQ3:} 
Can \shortname~ outperform other classification debiasing methods when used for data augmentation? Can it also mitigate bias in conditional diffusion models beyond text-to-image tasks?
\end{itemize}
\begin{figure}[b]
    \centering
     \includegraphics[width=0.48\textwidth]{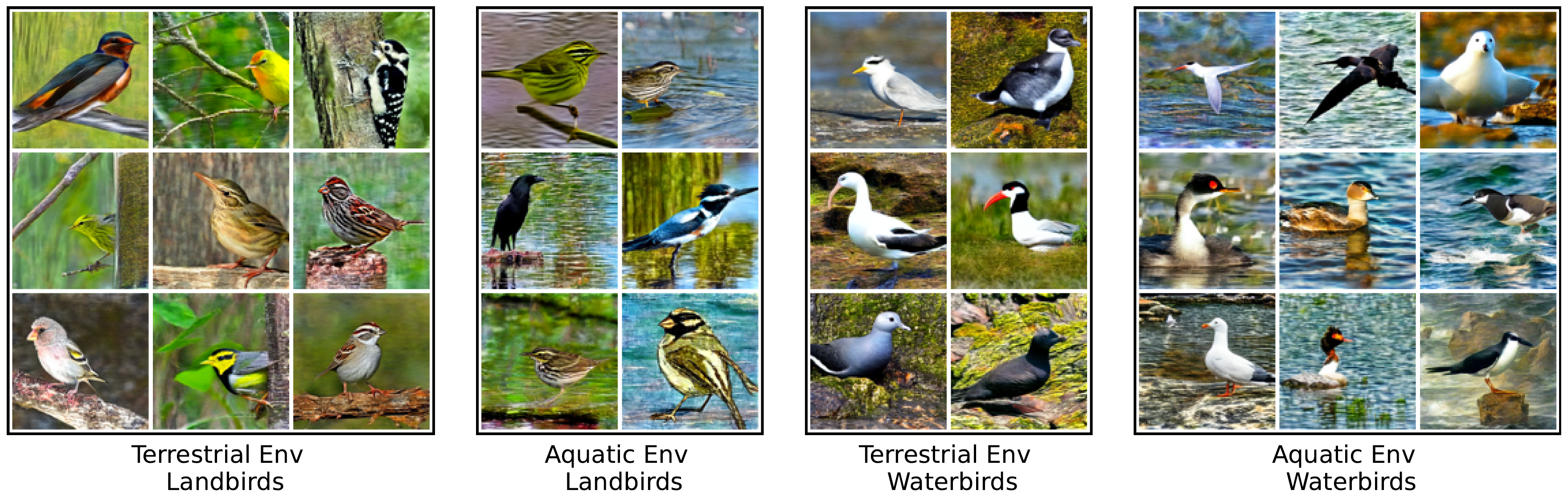}
    \caption{Images sampled from our unbiased model.}
    \label{fig:exp_samples}
    \Description{}
\end{figure}


\subsection{Experimental Settings}
\subsubsection{\textbf{Datasets.}} We conduct experiments on three publicly available benchmark datasets. Note that we access the bias attribute annotation only for the data construction and evaluations. The training sets are biased, while the test sets are not. 

\textbf{(1) Waterbirds}\cite{Sagawa*2020Distributionally,yao2022improving}: The Waterbirds image dataset contains two major categories of birds: waterbird and landbird, and each category has several specific species of birds. Images are spuriously associated with the background "water" or "land". There are 4,795 training samples while only 56 samples are “waterbirds on land” and 184 samples are “landbirds on water”. The remaining training data include 3,498 samples from “landbirds on land”, and 1,057 samples from “waterbirds on water”.
For the prompt settings, because there are significant morphological differences between different bird species, even if both species
are waterbirds or landbirds, there are still substantial differences
between them. To be more realistic, we set the prompts during
training and testing to specific bird species. 

\textbf{(2) CelebA}\cite{liu2015deep,Sagawa*2020Distributionally}: CelebA defines an image classification task where the input is a face image of celebrities and we use the classification label as its corresponding gender. We follow the data preprocess procedure from~\cite{Sagawa*2020Distributionally}. The label is spuriously correlated with hair color – “blond” or “black”. In CelebA, the minority groups are (blond, male) and (black, female). 
For more extensive testing, we constructed different group ratios for the following four groups: (blond, male), (blond, female), (black, male), (black, female). The sample ratio is 1:2:2:1. 
The (blond, male) group has 1,387 samples, which is the total number of blond male samples in the dataset.

\textbf{(3) FairFace}\cite{karkkainen2021fairface}: FairFace is a dataset balanced in terms of gender and race, using binary gender and including eight races. Considering the accuracy of the classifier during evaluation, we consolidate them into four broader classes, following previous work: WMELH = \{White, Middle Eastern, Latino Hispanic\}, Asian = \{East Asian, Southeast Asian\}, Indian, and Black. Our data consist of eight groups: (Female, White), (Female, Asian), (Female, Indian), (Female, Black), (Male, White), (Male, Asian), (Male, Indian), and (Male, Black). For unbiased data, the ratio is 1:1:1:1:1:1:1:1, and is 3:2:1:1:1:1:2:3 for biased data. The minimum group size is 1,500 samples. For the biased dataset, there are a total of 21,000 samples, while the unbiased dataset contains 12,000 samples.

\subsubsection{\textbf{Evaluation Metrics.}}
We select three types of metrics to evaluate the experimental results from different perspectives. 
\textbf{(1) Bias Metric}~\cite{shen2023finetuning}, which assesses the extent of bias in the results of the generative model. 
For every prompt \texttt{P}, we compute the $\operatorname{bias}(\texttt{P})=\frac{1}{K(K-1) / 2} \sum_{i, j \in[K]: i<j}|\operatorname{freq}(i)-\operatorname{freq}(j)|$, where freq($i$) is class $i$'s frequency in the generated images.
We train the environment classifier for the Waterbirds Dataset, hair color classifier for the CelebA Dataset, and the race classifier for the FairFace dataset. The number of class $K$ is 2/2/4, and the number of images for each prompt is 32/128/123 for Waterbirds/CelebA/FairFace.
\textbf{(2) Generation Quality Metric.} We use CLIP-T~\cite{shen2023finetuning}, the CLIP similarity between the generated image and the prompt, to evaluate the generation quality. We choose the CLIP-ViT-Base-16~\cite{pmlr-v139-radford21a} for evaluation.
\textbf{(3) Hybrid Metrics}: We use FID and Recall~\cite{kynkaanniemi2019improved} to measure the difference between the generated results and the original unbiased test data distribution. We used the VGG16~\cite{simonyan2014very} for evaluation.
\begin{figure}
    \centering
    \includegraphics[width=1\linewidth]{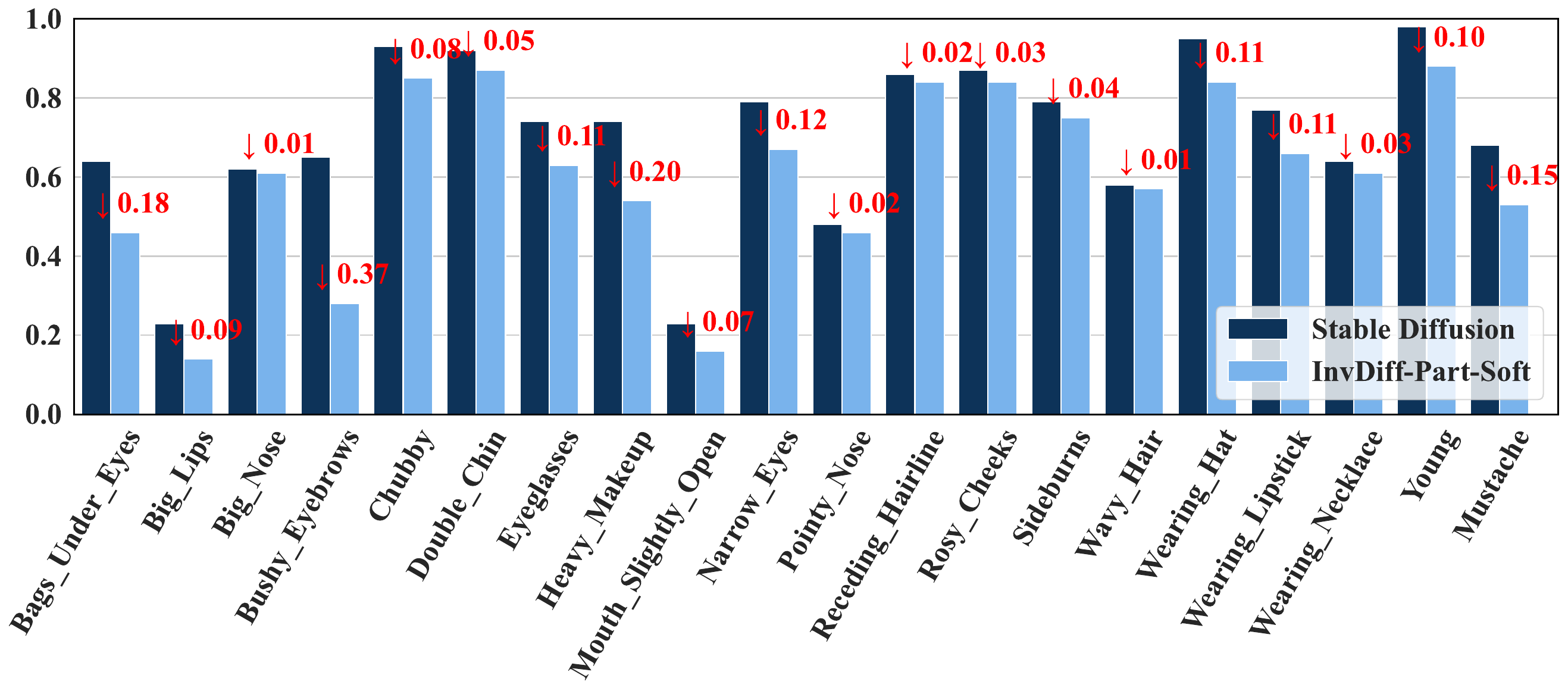}
    \caption{\shortname's results on multiple unknown biases.}
    \label{fig:unknown_biases}
\end{figure}

\subsubsection{\textbf{Comparison Methods.}}
The competitive baselines can be categorized into four groups.
\textbf{(1)} \textbf{The Stable Diffusion model} which is trained on the biased training dataset.
\textbf{(2)} SOTA diffusion debiasing methods: As far as we know, \shortname~ is the first study to mitigate unknown biases in diffusion models without relying on auxiliary bias annotations. Therefore, we choose two SOTA diffusion debiasing methods relying on auxiliary bias annotations for comparison.
\textbf{TIW}~\cite{kim2024training} is a time-dependent importance reweighting method designed to mitigate diffusion models' biases. TIW requires an unbiased dataset as annotations. \textbf{Fair-Diffusion}~\cite{shen2024finetuning} aims to reduce known biases associated with human faces, such as gender and race. It needs a bias classifier as an annotation and can't be used in the Waterbirds dataset.
\textbf{(3)} The third group contains four ablation counterparts of \shortname, \textbf{\shortname-Full-Hard}, \textbf{\shortname-Part-Hard}, \textbf{\shortname-Full-Soft}, and \textbf{\shortname-Part-Soft}. The "-Full" suffix denotes finetuning all the diffusion model's parameters. The "-Part" suffix denotes only training a small network $\boldsymbol{G}_\psi$ as Eq.(\ref{eq:final_loss}). The "-Hard" suffix represents the model use bias annotation in the dataset as the group result. The "-Soft" suffix denotes we obtain the group result using Eq.(\ref{eq:grouper_loss})
\textbf{(4)} We use the image generated by \shortname~as a data augmentation method to verify the debiasing effectiveness. We select vanilla \textbf{ERM}~\cite{vapnik1999overview} and SOTA debiasing classification methods without bias annotation for comparison, including \textbf{Mixup}\cite{zhang2018mixup}, \textbf{LfF}\cite{nam2020learning}, \textbf{Resample}~\cite{japkowicz2000class}, \textbf{Reweight}~\cite{japkowicz2000class}, \textbf{EIIL}~\cite{creager2021environment}.
(See Appendix \ref{appendix:experimental_details}  for the detailed training configurations).
\begin{figure*}[htbp]
    \begin{minipage}[b]{0.7\linewidth} 
        \centering
        \small

        \begin{tabular}{c|cccc|cccc}
            \toprule
            & \multicolumn{4}{c|}{$\Delta=0.2$} & \multicolumn{4}{c}{$\Delta=0.3$} \\  
            \midrule
            $\boldsymbol{G}_\psi$ Parameters& FID $\downarrow$& Rec $\uparrow$& Bias$\downarrow$& CLIP-T $\uparrow$& FID $\downarrow$& Rec $\uparrow$& Bias$\downarrow$& CLIP-T $\uparrow$\\ 
            \midrule
            860M(Param0)& \textbf{78.29}& \textbf{0.17}& 0.55(0.41)& 32.46(1.96)
            & \textbf{77.14}& \textbf{0.15}& 0.55(0.40)& 32.31(1.99)\\ 
            551M(Param1)& 79.02&0.13& 0.55(0.36)& 32.55(2.03)
            & 80.86& 0.14& 0.71(0.39)& 32.45(2.06)\\ 
            220M(Param2)& 82.54& 0.09& \textbf{0.50(0.46)}& 32.18(2.17)
            & 220.08& 0.00& \textbf{0.49(0.29)}& 23.14(2.65)\\
            56M(Param3)& 83.28& 0.16& 0.69(0.35)& \textbf{32.82(2.00)}
            & 85.02& 0.10& 0.66(0.30)&\textbf{32.82(2.07)}\\
            15M(Param4)& 84.37& 0.13& 0.71(0.33)& 32.75(2.04)& 78.3& 0.11& 0.83(0.34)&32.79(1.94)\\
            \bottomrule
        \end{tabular}
        \subcaption{Impact of $\boldsymbol{G}_\psi$ Parameters on \shortname's Performance.}
        \label{tab:tab_deltaparam}
    \end{minipage}
    \hspace{0.05\linewidth} 
    \centering
    \begin{minipage}[b]{0.2\linewidth} 
        \centering
        \includegraphics[height=0.13\textheight, width=0.9\textwidth]{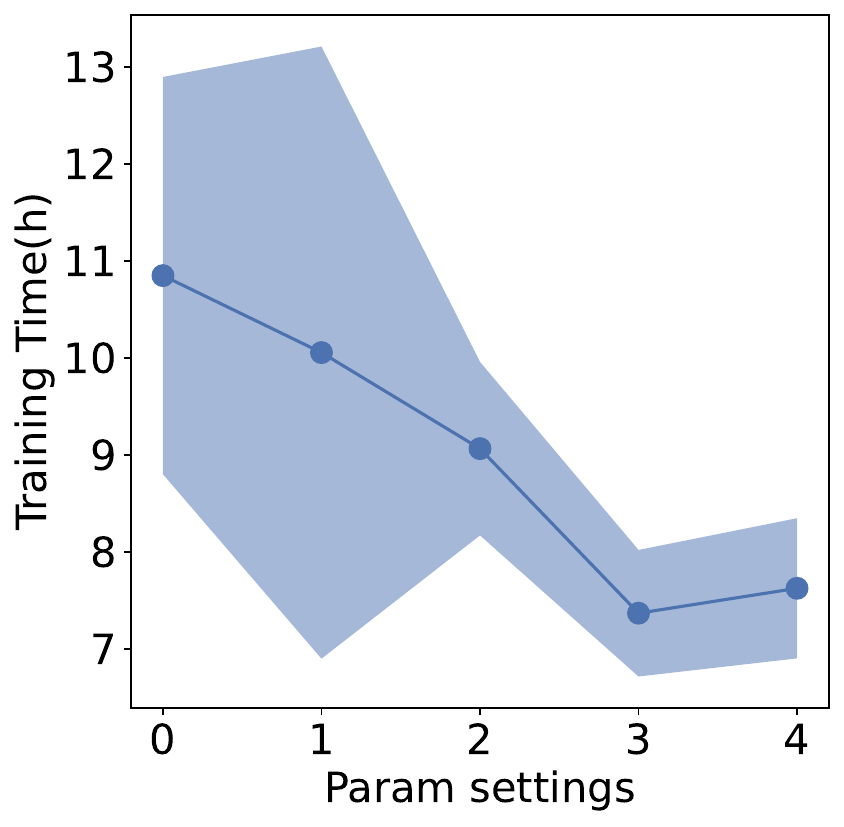}
        \subcaption{Training Time.}
        \label{fig:exp_training_time}
    \end{minipage}
    \caption{The Impact of $\boldsymbol{G}_\psi$ Parameters on Model Performance and Training Time.}
\end{figure*}

\subsection{Image Generation Results (RQ1)}
We first present the comparison between \shortname~ and the baseline methods across three datasets, as shown in Table \ref{tab:tab_main}. For the Bias and CLIP-T metrics, the value before the brackets represents the mean, while the value in brackets indicates the variance. The experimental results demonstrate that our method consistently achieves the lowest bias across all three datasets. With bias annotation, our model’s debiasing performance (\shortname-Full-Hard and \shortname-Part-Hard) significantly outperforms the state-of-the-art comparison method that also utilizes bias annotation. Even without bias annotation, our models (\shortname-Full-Soft and \shortname-Part-Soft) still produce superior results in most cases. Additionally, the FID, Recall, and CLIP-T values are comparable to those of the Stable Diffusion model, indicating that \shortname~ can maintain the quality of generated images while effectively reducing bias. Figure \ref{fig:exp_samples} shows the image randomly sampled from our unbiased model.

Nevertheless, we observe that when sensitive attribute annotations are unavailable, the debiasing effect of \shortname-Part-Soft on the CelebA dataset seems less pronounced~(Bias 0.80 $\rightarrow$ 0.70). The reason is that compared to other datasets (FairFace with only human faces, Waterbird with birds and two kinds of backgrounds), CelebA contains more complex features including hair, eyeglasses, hats, mustaches, etc. Therefore, there may be many latent biases in CelebA, while the bias constructed in the CelebA dataset is only between gender and hair color. Without bias annotations, the soft-version model likely has to balance not only the biases between hair color and gender but also other complex unknown biases. 
To validate this, we further conduct experiments on CelebA to explore whether our soft-version model can mitigate other unknown biases. Specifically, we first trained 20 binary classifiers for 20 face-related attributes using 200,000 images collected from the CelebFaces Attributes Dataset~\footnote{https://mmlab.ie.cuhk.edu.hk/projects/CelebA.html}. 
We then assessed whether the images generated by our models on CelebA can mitigate these potential biases in Stable Diffusion. The bias metric results in Figure \ref{fig:unknown_biases} validate our suppose, confirming the significance of its performance without bias annotations.

\begin{figure}[!htp]
    \centering
    \begin{subfigure}[t]{0.20\textwidth}
        \centering
        \includegraphics[width=1\textwidth,height=0.20\textheight]{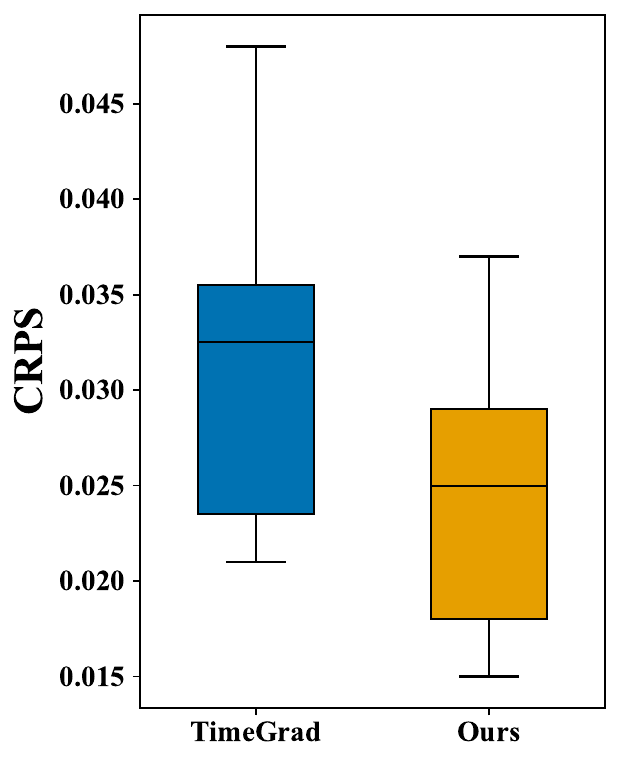}
        \caption{300 Epochs}
        \label{fig:ts_1}
    \end{subfigure}
    \begin{subfigure}[t]{0.20\textwidth}
\includegraphics[width=1\textwidth,height=0.20\textheight]{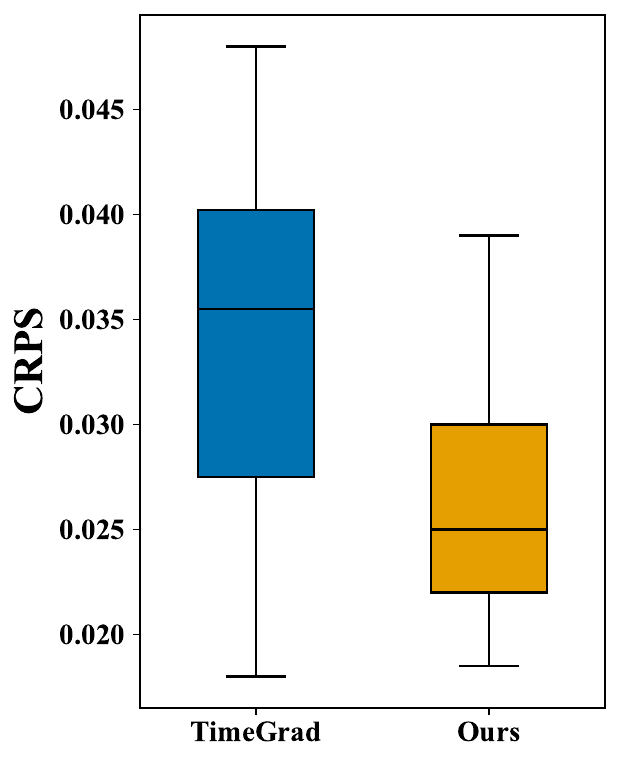}
        \centering
        \caption{500 Epochs}
        \label{fig:ts_2}
    \end{subfigure}
\caption{\shortname~ for Time Series Forecasting.}
\label{fig:ts}
\end{figure}


\subsection{Hyperparameter Sensitivity Analysis (RQ2)}
$\bullet$ \textbf{Impact of Parameter Quantity of $\boldsymbol{G}_\psi$.} In our settings, we fix the pretrained conditional diffusion model $\boldsymbol{\epsilon}_\theta$ and train the lightweight learnable module $\boldsymbol{G}_\psi$. We investigate the impact of parameter quantity of $\boldsymbol{G}_\psi$ and 
present the results for hyperparameter $\Delta = 0.2$ and 0.3 in Figure. \ref{tab:tab_deltaparam}. 
It shows that even when the parameters of $\boldsymbol{G}_\psi$ are only 15M, it can still achieve a certain degree of debiasing compared to the original biased model (Bias 0.89 $\rightarrow$ 0.71). However, when the parameter size of $\boldsymbol{G}_\psi$ is small, it may not ensure stable image generation quality. When the parameter size of $\boldsymbol{G}_\psi$ is around 200M (Param2), image quality begins to decline. Notably, when the parameter size of $\boldsymbol{G}_\psi$  ranges from 860M to 550M, the model maintains relatively stable output quality and debiasing capability.

\noindent $\bullet$ \textbf{Analysis of groups~(environments) number $E$.}
$E$ represents the number of groups we assume can potentially distinguish between various bias situations. In general, we can empirically choose $E$ as the product of the number of categories in sensitive attributes: $E=\prod_{i=1}^n C_i$, where $n$ is the total number of sensitive attributes considered (e.g., gender, hair color). $C_i$ is the number of categories in the $i$-th sensitive attribute. We found that our method remains relatively stable across all three datasets with respect to $E$. 
We performed a grid search between [2, 4, 8], and found that the performance was poor when set to 2, while both 4 and 8 yielded effective results. Ultimately, in our experiments, we set E=4, 4, 8 for Waterbird, CelebA, and FairFace, respectively.

\begin{figure}[!htp]
    \centering
    \begin{subfigure}[t]{0.23\textwidth}
        \centering
        \includegraphics[width=1\textwidth,height=0.16\textheight]{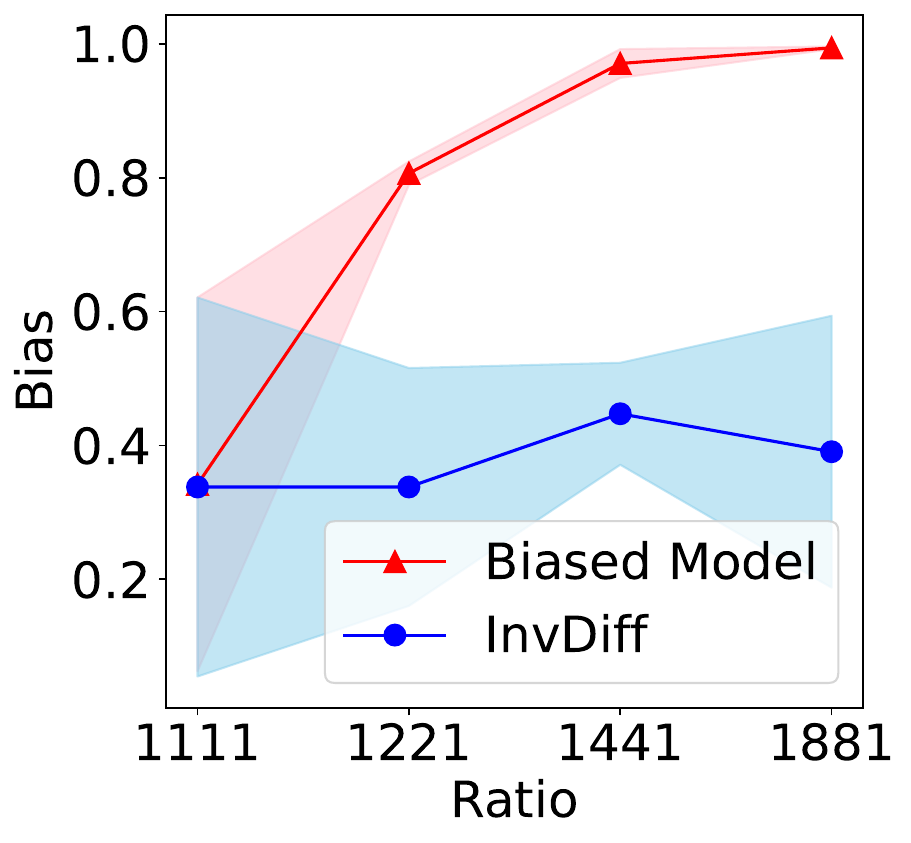}
        \caption{Multi-directional Bias}
        \label{fig:exp_data_ratio1}
    \end{subfigure}
    \begin{subfigure}[t]{0.23\textwidth}
\includegraphics[width=1\textwidth,height=0.16\textheight]{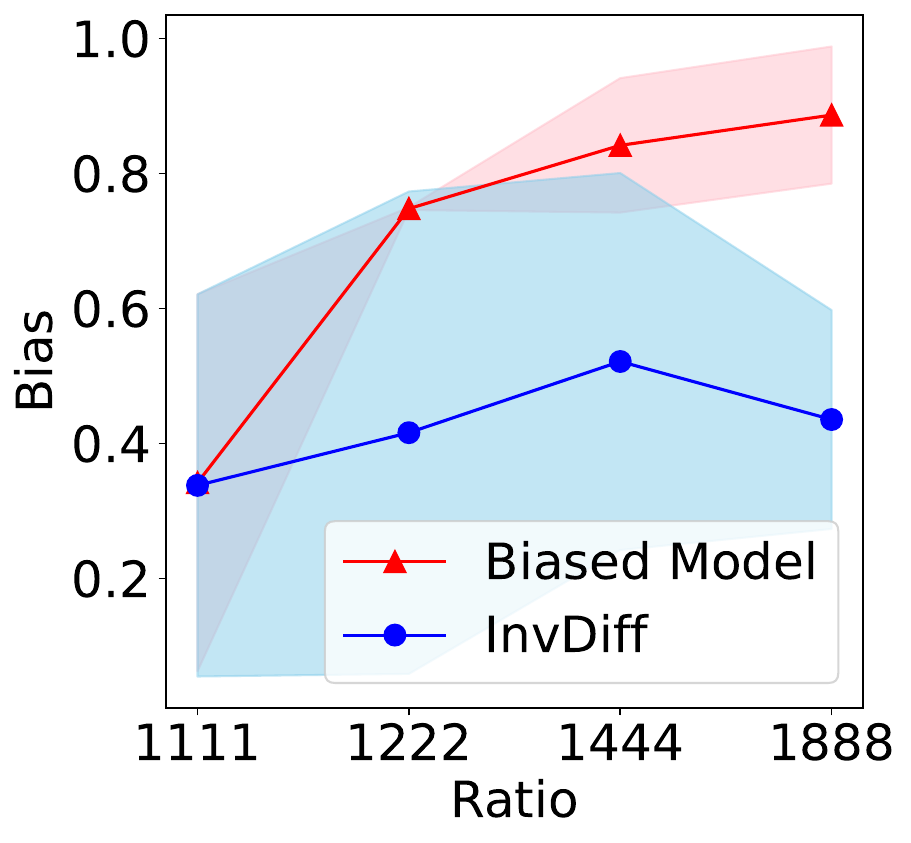}
        \centering
        \caption{Uni-directional Bias}
        \label{fig:exp_data_ratio2}
    \end{subfigure}
    \caption{Analysis of debiasing at different levels of bias in CelebA dataset.}
    \label{fig:exp_data_ratio12}
\end{figure}

\noindent $\bullet$
\textbf{Debiasing Results at Different Levels of Bias.}
In real-world data, the degree of bias is complex and uneven. Using the CelebA dataset as an example with two features, hair color and gender, the data might be biased in the blonde hair category while being unbiased in the black hair category, with varying degrees of bias. The model should be effective against different types of bias without making an unbiased model more biased.
Figure~\ref{fig:exp_data_ratio12} illustrates our model's debiasing capability for different types of bias. In Figure~\ref{fig:exp_data_ratio1}, the biased groups are blonde hair, black hair, blonde males, blonde females, black-haired males, and black-haired females, with proportions of 1:2:2:1, 1:4:4:1, and 1:8:8:1, respectively. In Figure~\ref{fig:exp_data_ratio2}, the proportions for the biased blonde hair and unbiased black hair groups are 1:2:2:2, 1:4:4:4, and 1:8:8:8. It can be seen that our model effectively addresses varying degrees of bias in the data without introducing bias in previously unbiased models (1:1:1:1) and can also correct imbalanced biases.

\noindent $\bullet$ \textbf{Training Time Analysis.}
When training our model, we train a module $\boldsymbol{G}_\psi$ to debias a biased model, where the parameter size of module $\boldsymbol{G}_\psi$ can be much smaller than that of the biased model.
We tested the training time for different parameter sizes of $\boldsymbol{G}_\psi$, as shown in Figure~\ref{fig:exp_training_time}. In these experiments, we used a single A100 GPU, FP16 mixed precision, 10,000 steps, and a batch size of 64. When the parameter size of the trainable module is reduced, the training time can be effectively decreased.

\noindent(See Appendix \ref{appendix:experiments} for more hyperparameter analysis).

\subsection{\shortname~ for More Tasks (RQ3)}
\noindent $\bullet$ \textbf{\shortname~ for Data Augmentation.} We further use \shortname~ as a data augmentation method, and compare its performance with SOTA debiasing methods that don't rely on bias annotation. The experimental results are shown in Table \ref{tab:augumentation}. For \shortname, we generate 4,795 samples and add them to the training set. \shortname~ demonstrates good performance, further illustrating the effectiveness of our proposed method.

\noindent $\bullet$ \textbf{\shortname~ for Time Series Forecasting.} 
We investigate whether \shortname~ can mitigate bias in conditional diffusion models beyond text-to-image tasks. 
We conduct additional experiments on time series forecasting tasks.
Out-of-distribution (OOD) time series data often exhibit similar bias characteristics to those found in generative models, where models tend to overfit irrelevant domain-specific factors. \shortname~ aims to address this by helping the model focus on the inherent patterns within the time series itself, rather than being influenced by extraneous, domain-specific variables. We conduct experiments on the OOD time series forecasting dataset AusElec~\cite{gagnon-audet2023woods}, and compare the performance against probabilistic time series forecasting backbone TimeGrad~\cite{rasul2021autoregressive}. 
For this task, the condition (input features) consists of previous electricity demand time series data. The generated result (target variable) corresponds to the future electricity demand. The dataset has 13 time domains, where each domain contains data from different months and holidays. We train the model on 12 domains and then test it on the remaining one at a time.
We utilized the widely used metric Continuous Ranked Probability Score (CRPS) to measure how good forecasts are in matching observed outcomes. The smaller of CRPS the better.
Figure \ref{fig:ts} shows the training dynamics for both \shortname~ and the biased backbone TimeGrad. As training progresses, \shortname~ continues to improve in both accuracy and consistency, with a noticeable reduction in distribution variance. This behavior is consistent with the model learning invariant temporal features, which are crucial for effective time series forecasting across domains. In contrast, the baseline model begins to show signs of overfitting, with accuracy plateauing and increased variance. This highlights that \shortname~ benefits from its ability to learn more robust, domain-agnostic features, enabling it to avoid overfitting while better adapting to diverse data distributions.
This study highlights that \shortname~ is effective not only in text-to-image generation but also in extending its debiasing capabilities to time series forecasting, showcasing its versatility across diverse applications.


\begin{table}
\small
\centering  
\caption{\shortname~for Data Augmentation}
\label{tab:abla_stu}  
    \begin{tabular}{cccccc}  
    
    \toprule  
    Method & Acc. $\uparrow$ & Balanced Acc. $\uparrow$ & AUROC $\uparrow$ & Worst Prec. $\uparrow$\\ \midrule  
    ERM & 0.841 & 0.831 & 0.910 & 0.607\\  
    Mixup & 0.892 & \textbf{0.889} & \textbf{0.949}& \underline{0.716}\\
    LfF &0.866 & \underline{0.863} & 0.934 & 0.647\\  
    Resample & 0.862 & 0.850 & 0.928 & 0.649\\  
    Reweight & 0.862 & 0.862 & \underline{0.940} & 0.661\\
    EIIL & \underline{0.901} & 0.879 & 0.939 & 0.746\\
    \midrule  
    \shortname & \textbf{0.902} & 0.858 & \underline{0.940} & \textbf{0.842}\\
    \bottomrule  
    \end{tabular}  
\label{tab:augumentation}
\end{table}
\section{Conclusion}
In this paper, we addressed the problem of mitigating unknown biases in diffusion models without relying on bias annotations or unbiased datasets.
We proposed \shortname, a general debiasing framework over pre-trained conditional diffusion models by incorporating invariant semantic information as guidance.
Specifically, we employed a lightweight trainable model that utilizes the invariant semantic information to guide the diffusion model’s sampling process toward unbiasing. Simultaneously, we design a novel diffusion training loss that automatically learns invariant semantic information.
In \shortname, we only need to learn a small number of parameters in the lightweight learnable module without changing the pre-trained diffusion model. Experimental results on various datasets and settings validate \shortname's notable benefits.

\section*{ACKNOWLEDGMENTS}
We extend our sincere gratitude to Bowen Deng for his significant contributions to the time series forecasting experiments in Section 5.4, conducted during the camera-ready phase. This work was supported in part by grants from the National Natural Science Foundation of China (Grant No. 62402159, U23B2031, 72188101), the National Key Research and Development Program of China (Grant No. 2021ZD0111802), and the Fundamental Research Funds for the Central Universities (Grant No. JZ2023HGQA0471, JZ2024HGTA0187).

\bibliographystyle{ACM-Reference-Format}
\bibliography{sample-base}

\appendix
\section{Appendix: Experimental Details}
\label{appendix:experimental_details}

\subsection{Training Configuration}
\label{sec:appendix_trianing_conf}
\subsubsection{Model Architecture}
Our method learns a parameter-efficient gradient estimator $\boldsymbol{G}_\psi$ on a pre-trained biased model. For the pre-trained biased model, we select "CompVis/stable-diffusion-v1-4"~\cite{rombach2022high} and fine-tune it on the biased dataset to facilitate validation. For the network architecture of $\Delta$, we choose a UNet. The down block types are "CrossAttnDownBlock2D", "CrossAttnDownBlock2D", "CrossAttnDownBlock2D", and "DownBlock2D". The mid block type is "UNetMidBlock2DCrossAttn". The up block types are "UpBlock2D", "CrossAttnUpBlock2D", "CrossAttnUpBlock2D", "CrossAttnUpBlock2D". For more information, please refer to Table~\ref{tab:parm_size_list}. The structure of the pre-trained unbiased model is consistent with the Param0 structure in Table~\ref{tab:parm_size_list}.

\begin{table}[!htbp]
    \centering
    \small
    \caption{$\boldsymbol{G}_\psi$ under different parameter quantity settings }
    \begin{tabular}{ccc}
        \toprule
         &   Parameter Quantity of $\boldsymbol{G}_\psi$& Block Out Channels\\
         \midrule
         Param0&  860M& (320, 640, 1280, 1280)\\
         Param1&  551M& (320, 640, 960, 960)\\
         Param2&  220M& (160, 320, 640, 640)\\
         Param3&  56M& (64, 160, 320, 320)\\
         Param4&  15M& (32, 64, 160, 160)\\
         \bottomrule
    \end{tabular}
    \label{tab:parm_size_list}
\end{table}

\subsubsection{Searched Parameters}
We completed the experiment on a single card GPU 80G-A100.
The hyperparameter search ranges for training the model are as follows. We tune the batch size for all the models in \{8, 32, 64\}. Learning rate is chosen from \{1e-4, 1e-5\}. The scheduler of learning rate is chosen from \{constant, linear, cosine, cosine with restarts, constant with warm up\}. The warm up step is tuned from \{0, 500, 1000\}. Parameter quantity of $\boldsymbol{G}_\psi$ is chosen from \{860M, 551M, 220M, 56M, 15M\}.
For the hyperparameter $\Delta$ and $\lambda$, the range is \{0.1, 0.2, 0.3, 0.4, 0.5, 0.6, 0.8, 0.9, 1, 2, 10\} and 
\{0.2, 0.4, 0.6, 0.8, 1, 2, 5, 10, 20, 50, 100\}, respectively.

\subsection{Prompt Details}
\label{sec:appendix_prompt_details}
\subsubsection{Waterbirds}
We do not list training occupations here due to their large quantity. The test occupations are \texttt{[`a baird sparrow', `a bay breasted warbler', `a black capped vireo', `a blue grosbeak',       `a boat tailed grackle',  `a bronzed cowbird',  `a california gull',     `an american redstart',
        `a baltimore oriole',  `a belted kingfisher',       `a black tern',                  `a blue headed vireo',   `a bobolink' ,            `a brown pelican' ,   `a canada warbler' ,      `an anna hummingbird',
        `a bank swallow', `a black and white warbler',  `a black throated blue warbler', `a blue jay', `a brandt cormorant',     `a brown thrasher',   `an acadian flycatcher', `a barn swallow', `a black billed cuckoo', 
        `a black throated sparrow', `a blue winged warbler', `a brewer blackbird', `a cactus wren', `an american goldfinch', `a bird']}
\subsubsection{CelebA} The training and testing prompts are \texttt{[`a face of a celebrity without blond hair', `a face of a celebrity with blond hair']}
\subsubsection{FairFace} The training and testing prompts are \texttt{[`a face of a female',
                    `a face of a male']}

\begin{figure}[!htbp]
    \centering
\includegraphics[width=0.40\textwidth]{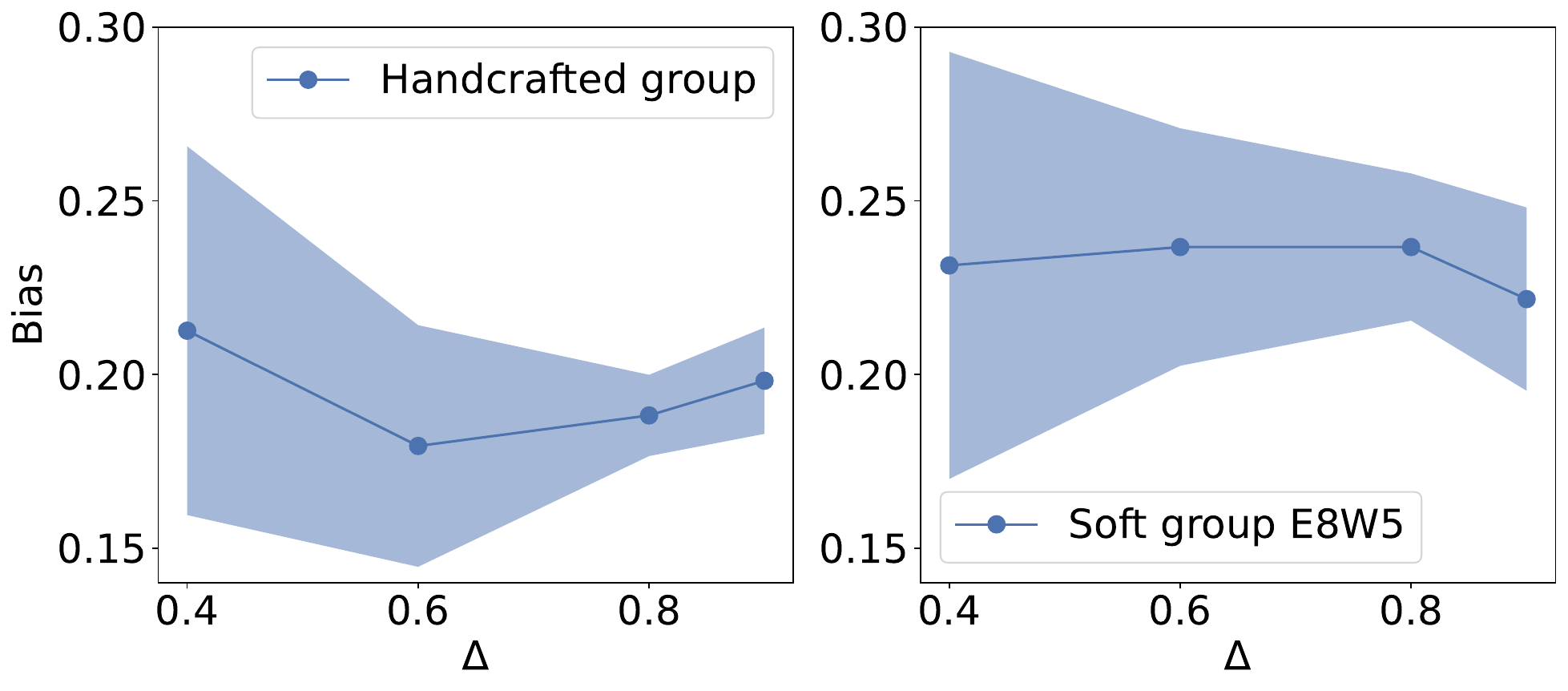}
    
    \caption{Changes in bias under different $\Delta$}
\label{fig:exp_delta_ratio_diff_grouper}
    \Description{}
\end{figure}
\begin{figure}[htbp]
    \centering
     \includegraphics[width=0.47\textwidth,height=0.47\textwidth]{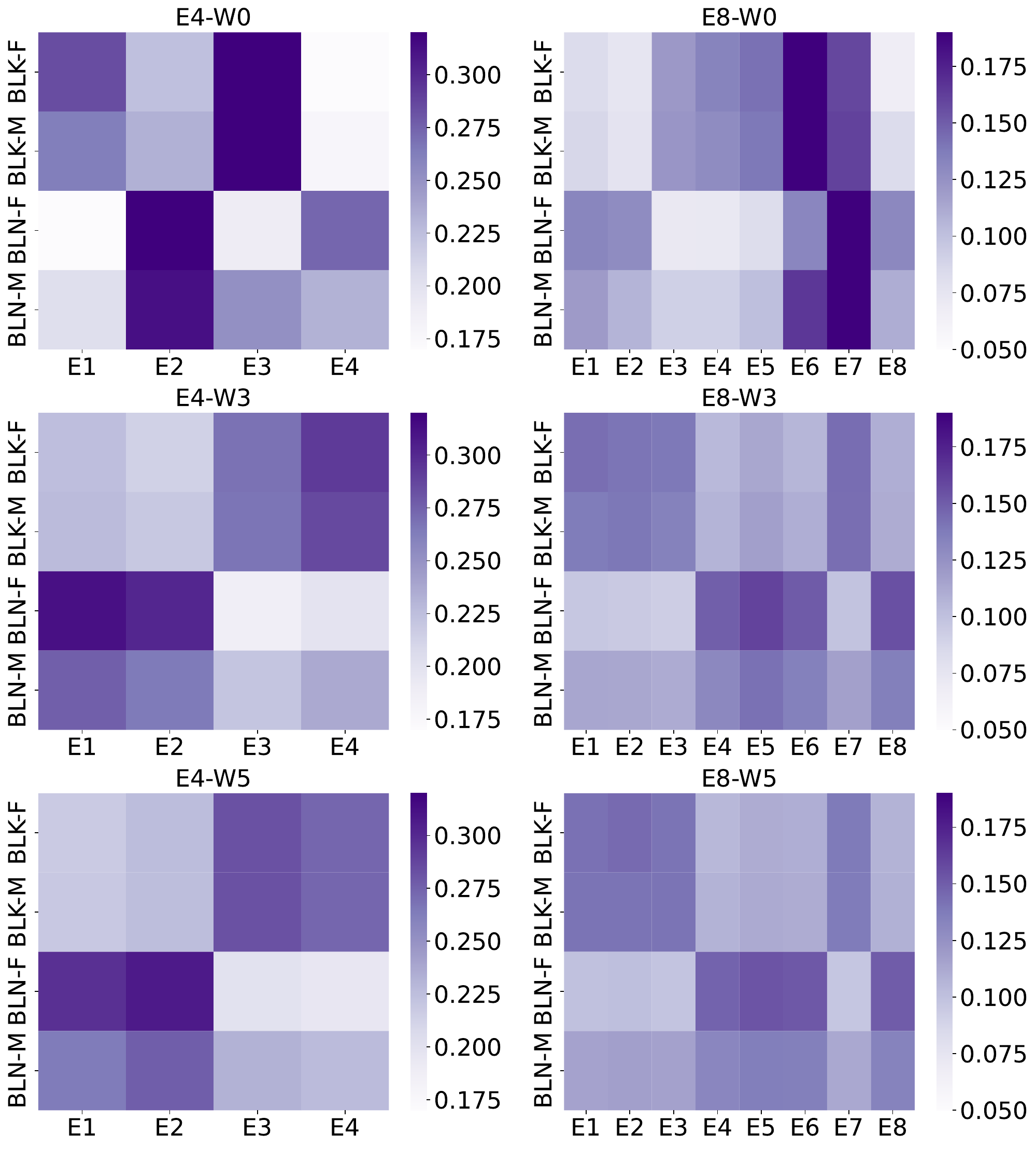}
    
    \caption{Correspondence between real features and soft grouper grouping. The figure shows the proportion of real data samples in each group, with each row summing to 1. Titles like Gx-Wy indicate x groups with parameter w set to y. Vertical axis labels are: BLN-M (blond hair, Male), BLN-F (blond hair, Female), BLK-M (black hair, Male), and BLK-F (black hair, Female). Horizontal axis labels, Gm, denote the m-th group, where the group number is arbitrary.}

    \label{fig:exp_grouper_heatmap}
    \Description{}
\end{figure}
\begin{figure}[!htbp]
    \centering
     \includegraphics[width=0.40\textwidth]{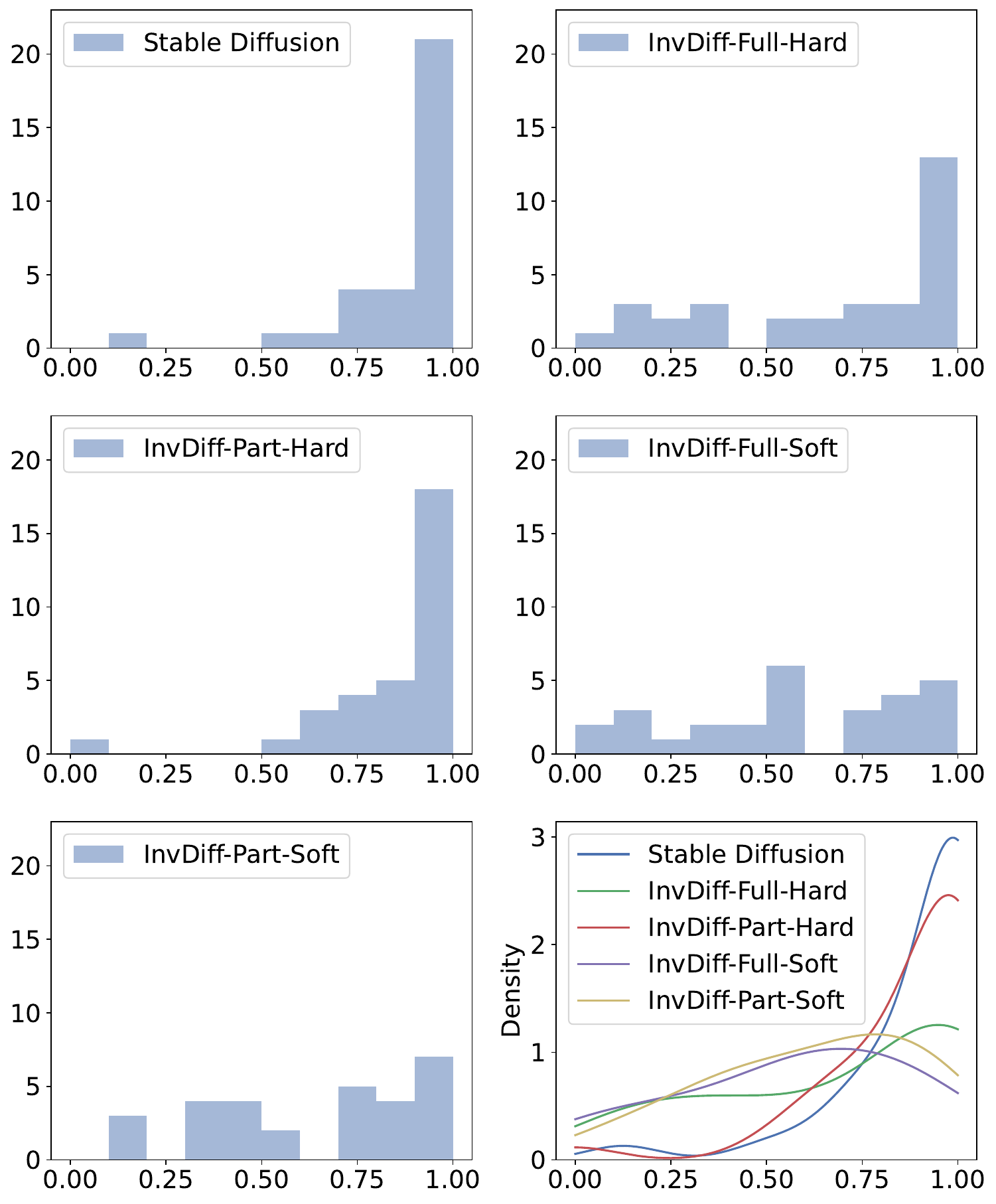}
    
    \caption{Distribution of multi-prompt debiasing results on Waterbirds dataset.}
    \label{fig:exp_bias_distribution}
    \Description{}
\end{figure}

\begin{figure*}[ht]
    \centering
     \includegraphics[width=0.8\linewidth]{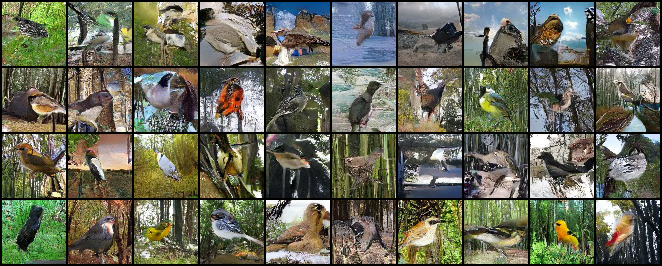}
    \caption{Samples from baseline TIW, trained on Waterbirds}
    \label{fig:samples_baseline_tiw_waterbird}
    \Description{}
\end{figure*}
\begin{figure}[htbp]
    \centering
     \includegraphics[width=0.40\textwidth]{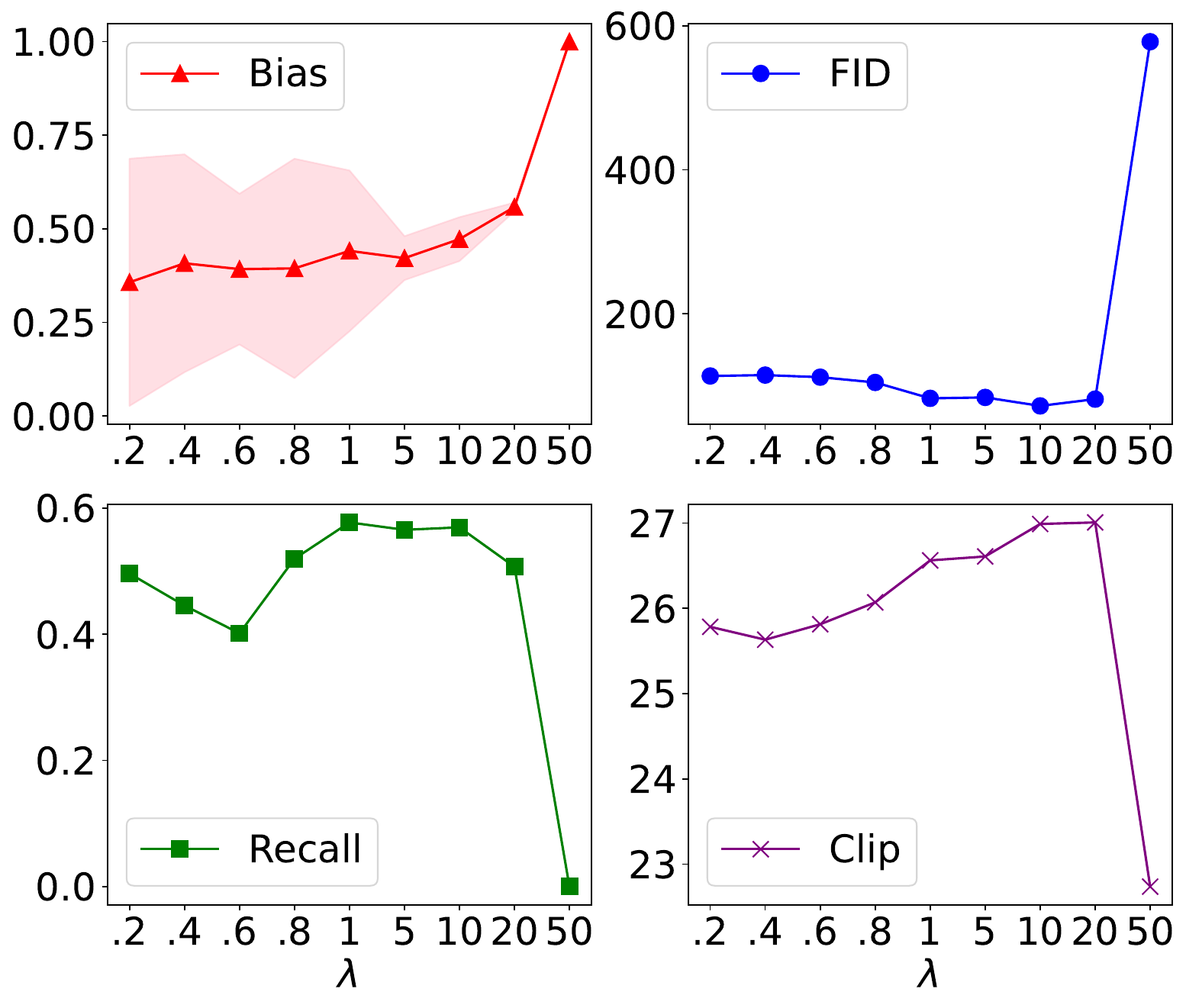}
    
    \caption{The effect of $\lambda$ on debias and image quality.}
    
    \label{fig:exp_lambda}
    \Description{}
\end{figure}

\begin{table*}[!htbp]
\small
    \caption{Impact of $\Delta$ under different groupers}
    \label{tab:tab_deltaratio_diff_grouper}
    \centering
    \begin{tabular}{c|cccc|cccc}
    \toprule
     & \multicolumn{4}{c|}{Handcrafted grouper} & \multicolumn{4}{c}{Softgrouper g8w5} \\  
     \midrule
     $\Delta$& FID $\downarrow$& Rec $\uparrow$& Bias$\downarrow$& CLIP-T $\uparrow$& FID $\downarrow$& Rec $\uparrow$& Bias$\downarrow$& CLIP-T $\uparrow$\\ 
     \midrule
     0.4& 132.01
& 0.59
& 0.21(0.05)& 25.8593(1.3508)
& 105.97
& 0.57
& 0.2314(0.0615)
& 25.9937(1.2988)
\\ 
     0.6& 134.97
&0.38
& 0.1794(0.0348)
& 25.9126(1.3115)
& 102.28
& 0.59
& 0.2367(0.0342)
& 26.1386(1.2346)
\\ 
     0.8& 134.84
& 0.66
& 0.1882(0.0117)
& 25.9088(1.3607)
& 87.74
& 0.54
& 0.2367(0.0212)
& 26.4237(1.2536)
\\
 0.9& 146.21& 0.74& 0.1982(0.0153)& 25.6917(1.3532)& 89.38& 0.5& 0.2217(0.0264)&26.6417(1.2410)\\
     \bottomrule
 
    \end{tabular}
\end{table*}.

\section{Experiments}\label{appendix:experiments}
\subsection{Hyperparameter Sensitivity Analysis}
\noindent $\bullet$ \textbf{Analysis of the dispersion degree $\omega$ of the grouping.}
When the dispersion degree $\omega$ is set larger, data with the same true label are more likely to be dispersed into different groups. To explore the mechanism by which the soft grouper functions, we tested our soft grouper under six settings on the CelebA dataset. In Figure~\ref{fig:exp_grouper_heatmap}, we show the distribution of data with different true features across different soft groups. It can be observed that when $\omega$ is small ($\omega = 0$), data with the same prompt are more likely to be grouped together, such as males and females with blond hair being concentrated in one group (for the four-group setting, this is $G2$; for the eight-group setting, this is $G7$). This makes it difficult for the model to distinguish biased features effectively. As $\omega$ increases, the distribution differences of samples with the same true features across soft groups become larger, which helps the model to achieve better debiasing.

\noindent $\bullet$ \textbf{Impact of $\lambda$.}
$\lambda$ is a hyperparameter that controls the degree of debiasing. The range of $\lambda$ is generally set from 0 to $\infty$. The larger the $\lambda$, the higher the degree of debiasing. We tested the effect of $\lambda$ values on model performance using the CelebA dataset, training on a biased dataset and testing on an unbiased dataset, see Figure~\ref{fig:exp_lambda}. Within the range of $\lambda = 0.2$ to $20$, we observed that as $\lambda$ increases, the mean of the bias shows a slow increase, but the standard deviation decreases significantly. This indicates that the debiasing effect is more stable and more robust to different prompts. The overall trend of FID decreases, and the overall trends of recall and CLIP-T increase, indicating that image quality is maintained or even improved. However, if $\lambda$ becomes too large, it can affect the quality of image generation. When $\lambda$ increases to 50, all metrics collapse significantly, indicating that the model cannot generate the target images effectively, let alone debias them.

\noindent $\bullet$ \textbf{Impact of $\Delta$}.
The $\Delta$ is a hyperparameter of the trainable model $\boldsymbol{G}_\psi$, controlling the extent of the debias module's influence on the original biased model. The parameter is set within the range of (0-1), with higher values indicating a greater influence on the original model. We tested the impact of the $\Delta$ on debiasing using the Fairface dataset. As shown in Figure~\ref{fig:exp_delta_ratio_diff_grouper}, results are presented for both the handcrafted grouper and the soft grouper. It can be observed that under both grouper settings, as delta increases, bias shows a decreasing trend, and the standard deviation range gradually narrows, indicating that a larger delta indeed facilitates debiasing. Additionally, we found that for the handcrafted grouper, a noticeable decrease in bias occurs when delta increases to 0.6, while for the soft grouper, a noticeable decrease is observed when delta increases to 0.9. This suggests that the optimal delta setting may differ for different types of groupers. More experimental results can be found in Table~\ref{tab:tab_deltaparam}.

\noindent $\bullet$ \textbf{Analysis of multi-prompt debiasing on the Waterbirds dataset.}
We know that for generative models, the acceptable set of prompts is infinite, and the model needs to debias across different prompts. We tested 30 prompts on the Waterbird dataset, representing 30 different types of birds. Figure~\ref{fig:exp_bias_distribution} shows histograms and kernel density estimation plots of the bias distribution frequency under different settings. The x-axis represents the values of the bias metric, and the y-axis shows the number of prompts with bias values falling within the corresponding interval. It can be seen that for the biased model, the bias metric is concentrated in the range of 0.8-1.0, whereas for our model, the prompts falling within the 0.8-1.0 bias range have significantly decreased. Our model can effectively mitigate this bias.

\subsection{Samples}
Figure~\ref{fig:samples_baseline_tiw_waterbird} shows the generation results of baseline TIW after training on the Waterbirds dataset.


\end{document}